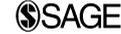

# Representation, learning, and planning algorithms for geometric task and motion planning

**Beomjoon Kim**[1] **, Luke Shimanuki**[2] **, Leslie Pack Kaelbling**[3]  **and Tomás Lozano-Pérez**[3]

**Abstract**

*We present a framework for learning to guide geometric task-and-motion planning (G-TAMP). G-TAMP is a subclass of task-and-motion planning in which the goal is to move multiple objects to target regions among movable obstacles. A standard graph search algorithm is not directly applicable, because G-TAMP problems involve hybrid search spaces and expensive action feasibility checks. To handle this, we introduce a novel planner that extends basic heuristic search with random sampling and a heuristic function that prioritizes feasibility checking on promising state–action pairs. The main drawback of such pure planners is that they lack the ability to learn from planning experience to improve their efficiency. We propose two learning algorithms to address this. The first is an algorithm for learning a rank function that guides the discrete task-level search, and the second is an algorithm for learning a sampler that guides the continuous motion-level search. We propose design principles for designing data-efficient algorithms for learning from planning experience and representations for effective generalization. We evaluate our framework in challenging G-TAMP problems, and show that we can improve both planning and data efficiency.*

**Keywords**
Task-and-motion planning, graph neural networks, generative adversarial networks, representations

## 1. Introduction

A mobile manipulation robot operating in complex environments such as construction sites or homes must solve *task-and-motion planning* (TAMP) problems. TAMP involves integrated task-level reasoning, such as deciding which object to move, and motion-level reasoning, such as deciding what motion to use to manipulate an object. TAMP is a difficult class of planning problems that involves a hybrid search space, long planning horizons, and expensive action feasibility checking.

One approach to solve TAMP problems is to take a pure-planning approach where a user designs a planner for the given problem (Cambon et al., 2009; Garrett et al., 2017, 2018; Gravot et al., 2005; Kaelbling and Lozano-Pérez, 2011; Srivastava et al., 2014; Toussaint, 2015). The advantage of this approach is its generality: for any given initial state and goal pair, most planners will eventually find a solution if there is one. The major drawback, however, is that it is computationally inefficient. TAMP planners do not typically have the ability to learn from past planning experience, and must solve difficult TAMP problems from scratch even when the current problem instance is similar to those solved in the past.

Alternatively, we can take a pure-learning approach to TAMP where we learn a policy that maps a state of the world to an action using reinforcement or imitation learning algorithms (Argall et al., 2009; Sutton and Barto, 1998). The key benefit is the computational efficiency: computing the next action to execute comes down to making a prediction using a function approximator, rather than performing an expensive search procedure. The downside, however, is its limited generalization capability. If a policy encounters a state that is very different from those seen in training, it may make mistakes and might get into a situation where it would not know how to proceed. Collecting a large amount

[1]Korea Advanced Institute of Science and Technology, Daejeon, Republic of Korea
[2]Optimus Ride, USA
[3]Computer Science and Artificial Intelligence Laboratory, Massachusetts Institute of Technology, Cambridge, MA, USA

**Corresponding author:**
Beomjoon Kim, Korea Advanced Institute of Science and Technology, Graduate School of Artificial Intelligence, 291 Daehak-ro, Daejeon, 34141, Republic of Korea.
Email: beomjoon.kim@kaist.ac.kr

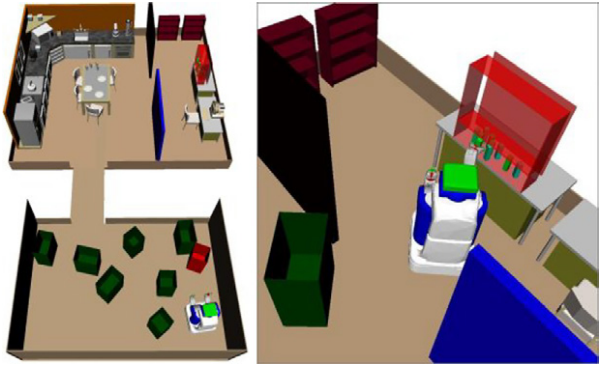

**Fig. 1.** Left: Moving the red box into the kitchen region. Right: Packing small objects inside the red shelf to the green box.

of data could be a solution, but data tends to be expensive in robot manipulation problems.

Based on these observations, we take the middle ground between these two extremes. We propose a framework that, given a set of planning experiences, learns to *guide* a planner by learning search guidance predictors. This approach has the best of both worlds. It is more efficient than pure planning because the predictors guide the search to more promising regions of the search space. In addition, it is far more robust than pure learning, as it can rely on the planner to correct for the mistakes the predictors may make.

We focus on a subclass of TAMP problems that we call *geometric task-and-motion planning* (G-TAMP), in which we are interested in moving a set of objects from one region to another among movable obstacles. This subclass is particularly important because it occurs in every TAMP problem, whether it is cooking a meal, constructing a building, or simply putting away groceries, a robot must efficiently reason about how to arrange obstacles to move objects to desired regions. Therefore, we believe if we can solve G-TAMP problems more efficiently, then we can solve TAMP problems more efficiently. Examples of G-TAMP problems are shown in Figure 1.

There are two key distinctions between a G-TAMP problem and standard graph search problem. The first is that G-TAMP problems involve a hybrid search space that consists of discrete task decisions and continuous motion decisions. The second is that G-TAMP involves expensive action feasibility checking: simulating a pick-and-place action, for example, requires a call to a motion planner and an inverse kinematics (IK) solver, making the generation of successor states expensive.

To deal with this, we propose a planning algorithm called *sampling-based abstract-edge heuristic search* (SAHS). Unlike traditional heuristic search, in which a state is expanded and its successors are added to the queue, SAHS maintains a priority queue of state-and-abstract-action pairs, which we call *abstract edges*. The queue is used to prioritize not only states, but also actions in order to perform feasibility checking on promising actions first.

Given a problem instance, SAHS searches forward, first branching on the choice of an *abstract action*, which specifies the type of a manipulation operator and its discrete parameters, such as $pick(obj_1)$. Then, given the choice of an abstract action, it branches on the continuous parameters of the operator, such as the grasp and base pose to pick the object, using random sampling. The planner calls a feasibility checker on the sampled parameters, and the next state is simulated. Figure 2 shows an example of the search tree of a G-TAMP problem.

The performance of SAHS heavily depends on the heuristic function for evaluating abstract edges and the sampler for sampling continuous parameters. One way to define them is by hand-designing them, but they would lack the ability to improve from planning experience. Instead, we develop learning algorithms that augment a hand-designed heuristic function and sampler.

There are two fundamental questions in designing these learning algorithms: what is a data-efficient objective function for learning from planning experience, and what is a representation that can generalize aggressively for G-TAMP problems. We describe our design principles for addressing these challenges.

## 1.1. Objective function design

A planning experience dataset consists of a set of trajectories that led to a goal, each of which is a sequence of state and action pairs. Like AlphaGo (Silver et al., 2016), one way to guide search is by learning the action-value function and prioritizing the search node expansion based on the action values. In our setting, we found this to be data-inefficient. Unlike a reinforcement learning setting in which you perform exploration to gather additional data, we have a fixed set of planning experience in which we only have the target values of the actions taken in the past plan solutions. So to learn the accurate value estimates for *all* the actions in a state, we would need to collect a large amount of data.

Instead, we design our objective function by being pessimistic about actions *not* taken in the states in our data. Given a state, the function learned this way would effectively suggest an action based on how frequently a similar state-and-action pair appeared in successful plans. In contrast to the approach that estimates action values, our strategy requires only the actions that achieved the goal.

Based on this principle for guiding the task-level planning, we propose to use the *large-margin* objective function (Tsochantaridis et al., 2005) to learn a ranking function for abstract actions. Given a state, a ranking function outputs rank values for the abstract actions, where the ranking among the actions is determined by their rank values. Given a state and action pair from our data, the objective function tries to maximize the margin between the rank value of the given action and the rest, such that the former has the highest rank.

We use the same pessimism principle to design the objective function for learning a sampler that guides the motion-level planning. More concretely, we use generative adversarial networks (GANS) (Goodfellow et al., 2014). Similar to the large-margin objective, in adversarial training, we learn a *discriminator* that assigns high values to the actions that occur in the dataset, and low values to the rest of the actions. We then train a *generator*, which in our case is the sampler, by tuning its parameters to generate actions that maximize the discriminator values. The result is a sampler that imitates the actions in the data.

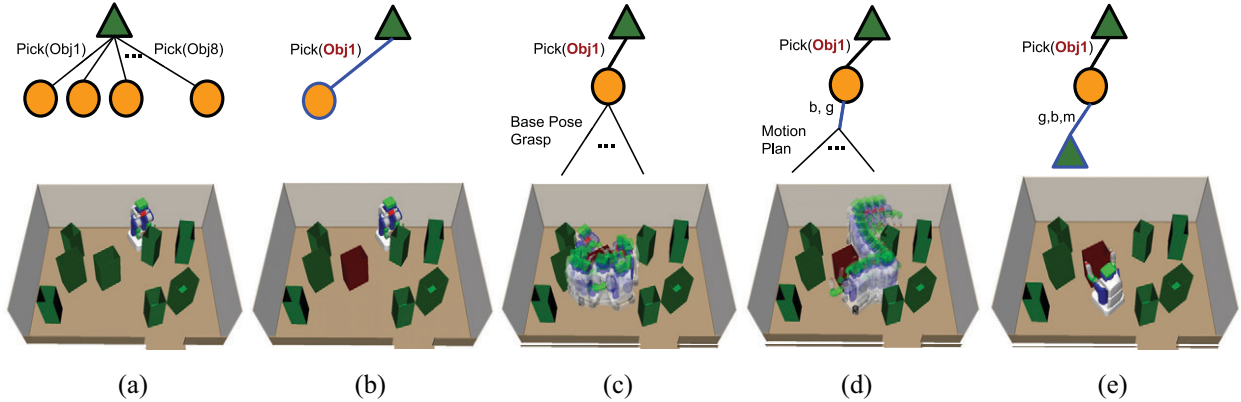

**Fig. 2.** A part of the search tree for SAHS. Green triangles denote nodes in which abstract action choices are available, and orange circles denote nodes in which continuous parameter choices are available. (a) The robot has several available abstract actions. (b) It chooses to explore abstract action Pick(obj₁) based on the given heuristic function. (c) The continuous parameters of the Pick action, which consists of base pose and grasp, are sampled using random sampling; ones that are in collision or do not have a feasible IK solution are rejected. (d) Once a feasible base pose *b* and grasp *g* are sampled, SAHS then calls a motion planner to see if there exists a motion plan to the selected base pose, and finds a feasible motion *m*. (e) The next state is generated, in which the robot has picked up the object with manipulation operation Pick(obj₁, (g,b,m)).

### 1.2. Problem representation design

A promising approach for guiding a search is to first represent the original planning problem in a relaxed form, solve it using a planner, and then use the solution to guide the search for the original problem. This representation design principle has been used in both task planning (Hoffman and Nebel, 2001) and motion planning (Schulman et al., 2014; Zucker et al., 2013). We use this intuition in our framework to design relaxed problem representations for G-TAMP problems.

Although both the ranking function and sampler use the representations designed based on this principle, we propose separate representations because we believe they fundamentally require different kinds of reasoning. The ranking function requires reasoning at the discrete task level, and the sampler requires reasoning at the continuous motion level.

For task-level planning, we relax a G-TAMP problem by representing the original problem abstractly, ignoring the geometric details of the environment, such as the poses and shapes of objects. More concretely, the abstract representation is defined as a conjunction of a novel set of *geometric predicates*. Each of these predicates computes reachability or occlusion information by calling motion planning algorithms. For instance, we have the OCCLUDESMANIP(o₁, o₂, r) predicate, which evaluates to true if object *o₁* is in the swept volume needed by the robot to manipulate object *o₂* into region *r*. Goals are also expressed using these predicates, by a conjunction of instances of the INREGION(o, r) predicate, each of which indicates that we wish to put object *o* in region *r*.

Using this representation, we propose a hand-designed heuristic function that recursively counts the remaining number of objects to move to get to the goal. To endow this function with the ability to improve with planning experience, we augment it using the abstract action ranking function. The problem, however, is that typical feed-forward neural networks (NNs) cannot use as an input the abstract problem representation that includes relational information and has varying numbers of objects across problems.

To deal with this, we first encode the abstract representation using a graph, where each node encodes predicates of an entity, and each edge encodes the predicates of a pair of entities. We then use graph neural networks (GNNS) which can take as an input a graph with varying sizes. Figure 3 (green) summarizes the flow of computations for computing the representation and the ranking among abstract actions.

The additional benefit of using such abstract representation is the immense generalization capability of the ranking network. Instead of having our ranking function perform reasoning based on the original problem representation that consists of the poses and shapes of the objects in the scene, we design it such that it only performs the high-level reasoning based on the abstract representation. The low-level geometric reasoning is delegated to motion planners that can be applied to a wide range of environments. This enables our ranking function to generalize to another environment with significantly different geometric details without retraining. We show that we can train our ranking function using only the training data collected in the environment shown in Figure 1(left), and it generalizes to the environment shown in Figure 1(right).

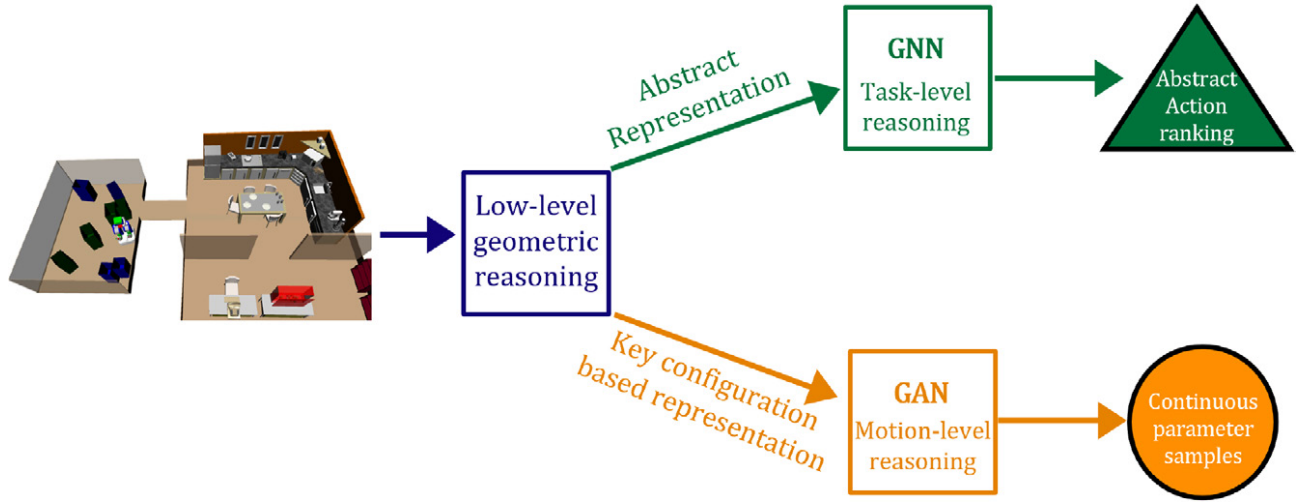

**Fig. 3.** Flow of computations for predicting an abstract action ranking (green) and continuous parameter sample (orange). Given a scene, a motion planning algorithm and collision-checker first performs low-level geometric reasoning to compute an abstract problem representation encoded in a graph, and key-configuration-based representation. For an abstract action ranking, the graph is fed to a graph neural network (GNN), which performs task-level reasoning to output a ranking. For continuous parameters, the key-configuration-based problem representation is fed to a general adversarial network (GAN), which performs motion-level reasoning to output parameter samples.

For motion-level planning, we first observe that the continuous parameters often specify, implicitly or explicitly, a goal configuration for a low-level motion planner. Thus, to predict promising continuous parameters, the learned sampler would have to verify the existence of a collision-free motion. Ideally, we would use the full configuration space (C-space) obstacles as input to the sampler, but this is difficult because we do not have an efficient way to compute these obstacles.

Instead, we relax the motion planning problem by approximating the C-space obstacles using a finite but important set of configurations. More concretely, we propose a novel C-space-based state-and-goal representation using *key configurations*. Key configurations are a sparsely sampled set of configurations that have been used in the past planning experience. Using the key configurations, we approximate the C-space obstacles using a binary vector indicating collisions at key configurations. We also encode the goal of the original G-TAMP problem using the swept volumes for moving the goal objects to their goal regions, which, if cleared, would achieve the goal. Figure 3(orange) summarizes the flow of computations for computing the representation and the continuous parameter samples.

We evaluate our planner, representation, and learning algorithms on two challenging G-TAMP problems shown in Figure 1. We show that using our design principle and combination of learning and planning, we achieve better planning and data efficiency compared with several benchmarks.

This work is an integrated and extended version of our prior papers (Kim et al., 2018, 2019a; Kim and Shimanuki, 2019). We make the following additional contributions.

- We illustrate how the task-level guidance learning proposed in Kim and Shimanuki (2019) and the motion-level guidance learning proposed in Kim et al. (2018, 2019a) can be integrated into a single framework. We conduct a set of experiments demonstrating that addition of each guidance to a planner leads to more efficient planning.
- We offer a unifying principle for designing the loss functions and representations for both motion and task level guidance learning. We perform a set of experiments comparing different representations and losses, and show that those that abide by our principle are the most effective.
- We propose a probabilistically complete version of our planning algorithm proposed in (Kim and Shimanuki, 2019).
- We use WGAN-GP (Gulrajani et al., 2017) that offers much more stable training for learning a continuous parameter sampler, in contrast to a standard GAN used in (Kim et al., 2018, 2019a).
- We substantially expand the description of the abstract problem representation, its encoding into a graph, and the GNN that uses that graph from (Kim and Shimanuki, 2019).

## 2. Related work

### 2.1. Learning to guide planning

The most famous system for learning to guide planning is AlphaGo Zero, which was developed for the game of Go (Silver et al., 2017). In AlphaGo Zero, Monte Carlo tree search (MCTS) is integrated with a value function and policy that are learned from past experience of playing Go. These learned predictors then guide MCTS into promising regions of the search space for the future instances of the game. Our framework can be seen as a version of AlphaGo Zero for G-TAMP problems.

Given the success of AlphaGo Zero, one may wonder if we can simply use the methods from AlphaGo Zero and apply them to G-TAMP problems. There are two key differences between G-TAMP and Go that make such direct application difficult. First, Go has a discrete action space where each action is placing a stone on a Go board. In contrast, G-TAMP problems have a hybrid action space and each action requires a feasibility check by typically calling a motion planner and an IK solver.

Another key difference is the variability in dimensionality of the state space. In Go, you may use a board image as a state representation across different instances of the game. In contrast, states of G-TAMP problems consist of objects and their attributes, and if you have different numbers of objects across different problem instances, then you have variable state dimensions. In addition, a single image of a scene is not a sufficient representation in general for G-TAMP problems.

Our approach for guiding the abstract action search is closely related to a substantial body of work on learning to guide task planning. One line of work learns a heuristic function on states to be used in planners based on heuristic search. This is typically formulated as supervised learning from planning experience on related problems. Perhaps the most successful of these methods (Yoon et al., 2006) learns domain-dependent corrections to an existing domain-independent heuristic. Many approaches learn how to best combine a variety of domain independent heuristics (Domshlak et al., 2010; Fink, 2007). Other approaches (Garrett et al., 2016), like ours, learn to rank actions directly. One of the algorithms in Pinto and Fern (2017), aims to learn an action-value function for ranking actions, which we found to be data efficient in our setup.

For learning to guide TAMP, Chitnis et al. (2016) proposed learning action-values. However, their search is over plan refinements, rather than abstract actions and they do not address the state representation problem. Driess et al. (2020a) proposed an approach that directly predicts a task plan from an initial image of the scene, based on which a motion-level planning was performed to find a motion plan that satisfies the predicted task plan. Our method differs in that (1) we assume we know the poses and shapes of objects, and (2) we provide guidance both at the task and motion levels based on a representation that can reason about occlusion, reachability, and collisions. Driess et al. (2020b) proposed to predict the feasibility of a discrete decision by learning a classifier using an initial image of the scene. Our sampler also learns to perform feasibility reasoning for generating continuous parameter samples, but using key-configuration-based representation that enables reachability reasoning; also, it not only generate feasible but also goal-achieving samples. Chitnis et al. (2020) proposed an approach that learns to drop state variables based on experience. This enables faster planning by allowing a TAMP planner to ignore a subset of objects.

There is a body of related work on guiding the choice of continuous parameters given the abstract action sequences (Chitnis et al., 2019; Kim et al., 2018, 2019a,b). In particular, we directly build on our previous work on using GANs for learning a sampler (Kim et al., 2018, 2019a), but in this work we use advanced GAN training technique that offers much more stable training (Gulrajani et al., 2017).

### 2.2. TAMP *algorithms*

There are several pure-planning algorithms for TAMP. Many approaches define separate strategies for planning abstract actions and finding continuous parameters. Typically, the abstract action planning (also know as task planning) is often done using a classical task planning algorithms (Helmert, 2006), whereas the continuous parameter search is performed either using sampling-based (Cambon et al., 2009; Garrett et al., 2017, 2018; Srivastava et al., 2014) or optimization-based (Toussaint, 2015) algorithms. A more extensive survey of methods can be found in Garrett et al. (2021).

One of the difficulties of using learning algorithms in these planners is that at the task-level, a planner only has access to an incomplete state where the low-level geometric details are not determined (we do not know the poses of objects). This makes it to difficult to predict the rank of abstract actions, which requires occlusion and reachability information. Similarly, for continuous parameters, it is difficult for the learned action sampler to suggest high-quality actions if the low-level geometric details are not known. For this reason, our planner, SAHS, performs a search with complete states.

G-TAMP problems are very closely related to manipulation among movable obstacles (Stilman et al., 2007). G-TAMP generalizes this class of problems by allowing moving multiple goal objects to goal regions, and lifting the assumption that the robot must touch each object once. Rearrangement planning (King et al., 2016; Krontiris and Bekris, 2015) can also be considered as a subclass of G-TAMP problem, where one is given goal object poses instead of goal regions.

### 2.3. *Graph neural networks*

For guiding the search for abstract actions, we use an abstract state and goal representation represented with a graph. To handle this graph-based input, we use GNNs. GNNs (Gori et al., 2005; Scarselli et al., 2009; Sperduti and Starita, 1997) (see surveys of Battaglia et al. (2018); Wu et al. (2019); Zhou et al. (2018)) incorporate a *relational inductive bias*: a set of entities and relations between them. In particular, we build on the framework of *message-passing neural networks* (MPNNs) Gilmer et al. (2017), similar to *graph convolutional networks* (GCNs) (Battaglia et al., 2018; Kipf and Welling, 2017). A key advantage of GNNs is that they learn a fixed-size set of parameters from problem instances with different numbers of entities. After learning, the GNN can be applied to arbitrarily large sets of objects and relations. This is crucial for G-TAMP problems where the number of objects varies widely across different problem instances.

### 2.4. Generative adversarial networks

We use GANs for training our biased-samplers for continuous parameters. Two recent advancements in generative-model learning, GANs (Goodfellow et al., 2014) and variational autoencoders (VAEs) (Kingma and Welling, 2014), are good choices for this purpose because an inference is simply a feed-forward pass through a network. For our purpose, which involve predicting poses of objects and robot, we require sharp samples because a small deviation could mean the difference between feasibility and infeasibility. GANs are known to generate sharp samples, albeit at the cost of potentially missing modes Li et al. (2017), and we use GANs to train our sampler.

The problem with the original GAN (Goodfellow et al., 2014), which minimizes Shannon-divergence, is its instability during training. WGANs (Arjovsky et al., 2017), which minimizes the Earth-mover's distance, have been shown to be more stable, but can lead to difficulty in optimization due to hard gradient clipping. We use WGAN with Gradient Penalty (WGAN-GP) (Gulrajani et al., 2017) which improves the training of WGAN by using a soft-enforced constraint on the gradient of the discriminator.

## 3. Planning problem setup

We assume that the environment of a G-TAMP problem consists of a set of fixed rigid objects $\mathbf{O}^{(F)} = \{o_i^{(F)}\}_{i=1}^{n_F}$, a set of movable rigid objects $\mathbf{O}^{(M)} = \{o_i^{(M)}\}_{i=1}^{n_M}$, and a set of workspace regions $\mathbf{R} = \{r_i\}_{i=1}^{n_R}$.

A state of the system is determined by the poses of the movable objects, each of which is denoted $P_{o_i^{(M)}}$, and the configuration $c \in \mathcal{C}$ of the robot. The poses of the objects and regions are defined relative to a *parent* object, which can be a movable object such as a tray, or a fixed object, such as the floor. We denote a state as $s \in \mathcal{S}$ where $s = (P_{o_1^{(M)}}, \ldots, P_{o_{n_1}^{(M)}}, c)$. All objects and regions have known and fixed shapes. We assume that states are fully observable.

We are given a set of $n_o$ manipulation operators, $\mathcal{O} = \{\mathfrak{a}_1, \ldots, \mathfrak{a}_{n_o}\}$. We assume that an operator manipulates an object to a region. Each operator takes in as inputs a fixed number of operation-specific discrete parameters $\delta \in \mathbf{O}^{(M)} \times \mathbf{R}$, which specify which object moves to which region. If the object is already in that region, then it moves it to a different pose in that region. The operator also takes in continuous parameters $\kappa \in K_\mathfrak{a}$. The robot might have one or more such operators available, such picking-and-placing, pushing, or throwing.

Each of the operators has a set of feasibility constraints, and we are given a set of external feasibility checkers that can check the feasibility of the given continuous parameters. Examples of such feasibility checkers are an IK solver and a motion planner.

We define the pair of operator type and its discrete parameters as an *abstract action*, which we denote with $\mathfrak{a}(\delta)$. For instance, an abstract action

$$\textsc{PickandPlace}(\delta = (obj_1, table_1))$$

specifies the robot moves $obj_1$ to $table_1$ using pick-and-place. An *action*, denoted $\mathfrak{a}(\delta, \kappa)$, is a concrete operation that can be executed by the robot. For instance, we may have

$$\textsc{PickandPlace}(\delta = (obj_1, table_1), \kappa = (g, p))$$

for pick-and-placing $obj_1$ to $table_1$ using the continuous parameter that consists of grasp $g$ and the placement pose $p$ on $table_1$. Based on $\kappa$, we call a motion planner and compute a motion plan associated with $\kappa$. Note that motion plan is not part of $\kappa$, mainly because we learn a generative model defined over $\kappa$ but not motions.

Each action $\mathfrak{a}(\delta, \kappa)$ induces a mapping $T(\,\cdot\,, \mathfrak{a}(\delta, \kappa))$ from a world state $s$, in which it is executed, to a resulting world state $s' \in \mathcal{S}$. If the operation is infeasible and cannot be legally executed in $s$, we let $s' = s$ or an absorbing "failure" state. To check the feasibility of an operation, we assume $T(\,\cdot\,, \cdot\,)$ calls external functions necessary to find a low-level motion. For example, for $\textsc{PickandPlace}, T(\,\cdot\,, \cdot\,)$ calls an IK solver to find the joint configuration for the given grasp, and a motion planner to find the motion plan to that joint configuration; if any one these cannot be found, then the operation is deemed infeasible. The transition function $T$ is deterministic. However, in practice, it may return different results for the same inputs depending on the result of the external function calls, which must compute the desired values within a time limit.

We specify a goal set $\mathcal{G}$ as a conjunction of statements of the form $\textsc{InRegion}(o, r)$, where $o \in \mathbf{O}^{(M)}$ and $r \in \mathbf{R}$, which are true if $o$ is contained entirely in region $r$. It is also possible to specify the final robot configuration as part of the goal for NAMO problems (Stilman and Kuffner, 2005), or final poses of objects for specifying rearrangement problems (King et al., 2016; Krontiris and Bekris, 2015).

A G-TAMP *planning problem* is characterized by $(\mathbf{O}^{(M)}, \mathbf{O}^{(F)}, \mathbf{R}, s_0, \mathcal{O}, \mathcal{G}, T)$, where $(\mathbf{O}^{(M)}, \mathbf{O}^{(F)}, \mathbf{R})$ defines the *environment*, and $s_0$ is the initial state. The objective is to find a sequence of actions that changes the state from $s_0$ to a state that satisfies $\mathcal{G}$.

## 4. Sampling-based graph search algorithm

SAHS is a greedy-search algorithm that prioritizes exploration of *abstract edges*, defined by a pair of a state and abstract action, using a heuristic function. Before we describe SAHS and our heuristic function, it is worthwhile to examine the characteristics of the task-level reasoning in G-TAMP problems to illustrate what we wish to capture in the heuristic function.

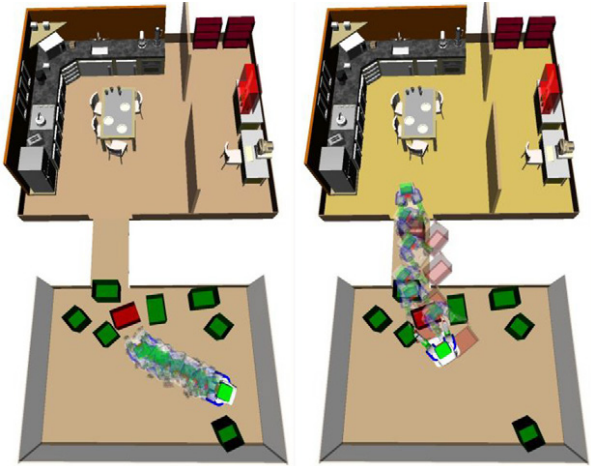

**Fig. 4.** Left: $V_{\text{pre}}(q, \mathfrak{a}(\delta, \kappa))$ where $\mathfrak{a}$ is pick-and-place, $\delta$ is the red box, $q$ is configuration of the solid robot; $\kappa$ is not shown. Right: $V_{\text{manip}}(\mathfrak{a}(\delta, \kappa))$ where $\mathfrak{a}$ is pick-and-place, $\delta$ is the red box and home region marked yellow, and $\kappa$ is the robot's base pose at the end of the trajectory inside the home region

In a G-TAMP problem, the given goal is achieved if we move each goal object to its corresponding goal region. Thus, in a state, the remaining number of objects to move to reach the goal is a useful heuristic. To enable computing such a heuristic, we construct an abstract problem representation that enables us to estimate it using recursive counting. Subsequently, we show how to encode this representation in a graph that can be used by a GNN to effectively suggest abstract action ranking.

### 4.1. Abstract problem representation

To capture reachability and occlusion information in a state, we make use of the assumption that each action manipulates a single object to move it to a region or to a different pose within its current region. Under this assumption, we can characterize the preconditions of executing an action $\mathfrak{a}(\delta, \kappa)$ with two volumes of workspace.

Assume $\delta = (o, r)$ for some object $o$ and for some region $r$, and the continuous parameters $\kappa$ are chosen so that if the robot uses action $\mathfrak{a}(\delta, \kappa)$, then it can move $o$ to region $r$. The first volume, $V_{\text{pre}}(q, \mathfrak{a}(\delta, \kappa))$, is the swept volume that the robot may move through from its current configuration $q$ to a configuration $q'$ in which object $o$ can be reached. The second volume, $V_{\text{manip}}(\mathfrak{a}(\delta, \kappa))$, is the swept volume that the robot and object may move through, from their configurations at the beginning of the operation, $(P_o, q')$, to their configurations at the end of the operation, as determined by continuous parameters $\kappa$. Figure 4 shows examples of these two swept volumes.

Of course, there are many such swept volumes that would suffice for each of these actions, corresponding to different choices of paths. We simply use a "nominal" path generated by a call to our path planner, which we describe shortly.

We construct a relational abstract representation of the state $s$ and goal $\mathcal{G}$, denoted $\alpha(s, \mathcal{G})$, as a conjunction of all true instances of the following predicates, applied to *entities* $e \in \mathbf{O}^{(M)} \cup \mathbf{R}$ in the environment:

- IsREGION($e$), true if $e$ is a region;
- IsOBJECT($e$), true if $e$ is an object;
- IsGOAL($e$), true if $e$ is mentioned in the goal specification $\mathcal{G}$;
- INREGION($o, r$), true if object $o$ is currently in region $r$;
- PREFREE($o$), true if $\exists \kappa$ such that $V_{\text{pre}}(q, \mathfrak{a}((o, r), \kappa))$ is collision-free;
- MANIPFREE($o, r$), true if $\exists \kappa$ such that $V_{\text{manip}}(\mathfrak{a}((o, r), \kappa))$ is collision free;
- OCCLUDESPRE($o_1, o_2$), true if $o_1$ is an object that overlaps the swept volume $V_{\text{pre}}(q, \mathfrak{a}((o_2, r), \kappa))$, where $\kappa$ is chosen to avoid collisions if possible; and
- OCCLUDESMANIP($o_1, o_2, r$), true if $o_1$ is an object that overlaps the swept volume $V_{\text{manip}}(\mathfrak{a}((o_2, r), \kappa))$, where $\kappa$ is chosen to avoid collisions if possible.

Example evaluations of OCCLUDESPRE and OCCLUDESMANIP are shown in Figure 5. The detailed implementations for the last four relations are specific to an operator *type*, such as pick-and-place or pull. The value of any of these predicates, if applied to arguments that are clearly the wrong type, is false.

Given state $s$ and goal $\mathcal{G}$, we must compute values for all instances of these predicates in that domain. The last four require non-trivial computation, including finding feasible $\kappa$ values and computing the motion plans to obtain the necessary swept volumes. This involves finding $\kappa$ and associated trajectories so that $V_{\text{manip}}$ and $V_{\text{pre}}$ have the minimum number of collisions with obstacles in the world. This problem, known as the *minimum constraint removal* (MCR) problem, is known to be very costly in the general case (Hauser, 2014).

Thus, we compute them by selectively ignoring movable objects. First, we attempt to find a collision-free $\kappa$ and trajectory. If that fails, then we simply find $\kappa$ and a trajectory that are collision-free with respect to the fixed obstacles, but may collide with movable obstacles. We also reuse samples of $\kappa$ and paths by extensively caching them to make the repeated computation of the predicates efficient. This is described in the appendix.

### 4.2. A default heuristic function for G-TAMP

We now describe our hand-designed heuristic function which estimates the number of objects that need to be moved to achieve the goal given the state representation $\alpha(s, \mathcal{G})$ and the abstract action $\mathfrak{a}(\delta)$. Recall that it must, in general, move the objects specified in the goal, as well as whatever other objects obstruct its ability to move those named in the goal. Let "M" denote a set of objects such that there is a way to move just those objects to achieve the goal, and such set with the minimum cardinality as $\mathbf{M}^*$. Ideally, we would define our heuristic function so that it maps $\alpha(s, \mathcal{G})$ to $|\mathbf{M}^*|$, which would be an optimal heuristic function. Unfortunately, determining $|\mathbf{M}^*|$ is NP-hard.

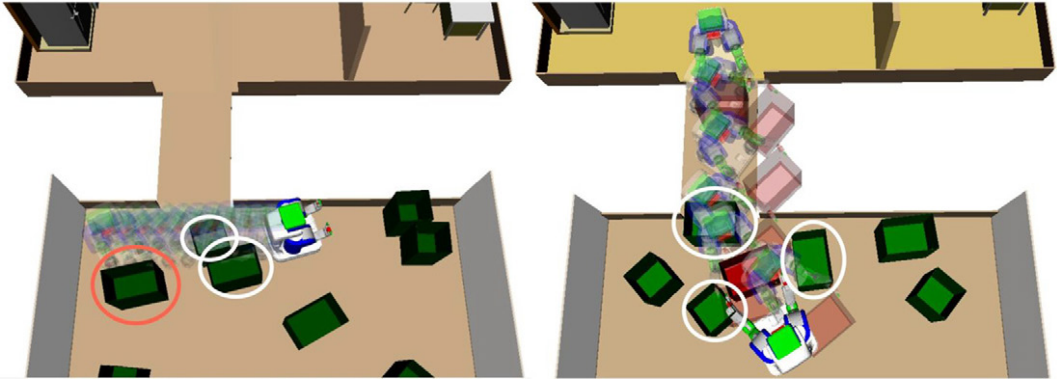

**Fig. 5.** Example evaluations of OccludesPre (left) and OccludesManip (right). In the left figure, the robot is trying to pick up the box marked with the red circle, and plans a motion whose swept-volume is shown with the transparent robots. There are two obstacles, marked with white circles, in collision with this motion, making OccludesPre($o_{white_1}$, $o_{red}$) and OccludesPre($o_{white_2}$, $o_{red}$) to true, and OccludesPre($o$, $o_{red}$) to false for all the other movable objects $o \in \mathbf{O}^{(M)}$. In the right figure, the robot is trying to move the red box to the kitchen region located on the top, and plans the manipulation motion whose swept-volume is shown with the transparent robots. There are three objects, marked with white circles, in collision with this motion, so OccludesManip($o_{white_i}$, $o_{red}$, $r_{kitchen}$) is true for $i \in \{1, 2, 3\}$, and OccludesManip($o$, $o_{red}$, $r_{kitchen}$) is false for all the other movable objects $o \in \mathbf{O}^{(M)}$.

Instead, we take a greedy approach. Our method estimates a set of objects the robot needs to move, M, starting with the goal objects. It then recursively considers objects that occludes the objects already in M, which are added to M. Our heuristic function based on this scheme, H-Count, is described in Algorithm 1.

The algorithm begins by initializing the set M with the goal objects not in their designated goal regions. It then initializes a queue with all the objects in M. Then, it pops an object $o_m$ from the queue, and checks whether for all $o \in \mathbf{O}^{(M)}$, OccludesPre($o$, $o_m$) is true or if OccludesManip($o_m$, $o$, $r$) is true. If so, $o$ is added to M and the queue. The process repeats until there are no more objects to be added.

H-Count is not an admissible heuristic. It overestimates $|\mathbf{M}^*|$, because the cardinality of M depends on the order in which we add objects to M. We greedily add objects that occlude objects already in M, instead of finding the ordering that would result in $|\mathbf{M}^*|$. Moreover, because our geometric predicates are not computed based on an exact MCR algorithm, the swept-volumes used for computing occlusions also tend to report more collisions than necessary.

Because this is an overestimate of the true cost-to-go, $|\mathbf{M}^*|$, we try to compensate by subtracting $|O_{achieved}(\alpha(s, \mathcal{G}))|$, which is the number of goal objects already in their designated goal regions in the given state; this would prefer states with more goal objects in their goal regions.

We also prioritize abstract edges (state-and-abstract-action pairs) instead of just states in order to prioritize feasibility checking on actions and expansions. To do this, we

---

**Algorithm 1** H-COUNT ($\alpha(s, \mathcal{G})$)

1:    $\mathbf{M} = \{o_G | (o_G, r_G) \in \mathcal{G} \land \neg \text{INREGION}(o_G, r_G)\}$
2:    $queue = Queue()$
3:    $queue.add(o)$ **for** $o \in \mathbf{M}$
4:    **while** $queue$ is not empty
5:      $o_m = queue.pop()$
6:      **for** $o \in \mathbf{O}^{(M)}$
7:        **if** OccludesPre($o$, $o_m$) $\lor \exists r$ s.t OccludesManip($o_m$, $o$, $r$)
8:          $\mathbf{M} = \mathbf{M} \cup \{o\}$
9:          $queue.add(o)$
10:      **end if**
11:     **end for**
12:    **end while**
13:    **return** $|\mathbf{M}|$

---

add a term that discourages manipulating goal objects that are already in the goal region. Putting everything together, our heuristic function for an abstract edge, which consists of an abstract state $\alpha(s, \mathcal{G})$ and abstract action $\mathfrak{a}(\delta)$, where $\delta = (o, r)$, is given by

$$H(\alpha(s, \mathcal{G}), \mathfrak{a}(\delta); \mathcal{G}) = \text{H} - \text{COUNT}(\alpha(s, \mathcal{G}))$$
$$- |O_{achieved}(\alpha(s), \mathcal{G})| \qquad (1)$$
$$+ \mathbb{1}_{\text{INREGION}(o, r) \land (o, r) \in \mathcal{G}}$$

### 4.3. Sampling-based abstract-edge heuristic search

Algorithm 2 describes SAHS, which takes as inputs an initial state $s_0$, set of goal states $\mathcal{G}$, hyperparameters for sampling continuous parameters of operators $N_{mp}$, and $N_{smpl}$, an abstract-edge heuristic function $\mathbf{h}(\cdot, \cdot)$, and *budget*, which

---

**Algorithm 2** SAHS $(s_0, \mathcal{G}, N_{smpl}, N_{mp,\,budget}, \mathbf{h}(\,\cdot\,,\,\cdot\,))$

---

1: **while** not budget reached
2:    $queue = \text{PriorityQueue}()$
3:    **for** $\delta \in \mathbf{O}^{(M)} \times \mathbf{R}, \mathfrak{a} \in \mathcal{A}$
4:       $queue.add((s_0, \mathfrak{a}(\delta)), \mathbf{h}(\alpha(s, \mathcal{G}), \mathfrak{a}(\delta)))$
5:    **end for**
6:    **while** not *queue.empty*
7:       $s, \mathfrak{a}(\delta) = queue.pop()$
8:       $\kappa = \text{SMPLCONT } (s, \mathfrak{a}(\delta), N_{smpl}, N_{mp})$
9:       **if** $\kappa$ is feasible
10:          $s' = T(s, \mathfrak{a}(\delta, \kappa))$
11:          $s'.plan = s.plan + \mathfrak{a}(\delta, \kappa)$
12:          **if** $s' \in \mathcal{G}$
13:             return $s'.plan$
14:          **end if**
15:          **for** $\delta' \in \mathbf{O}^{(M)} \times \mathbf{R}, \mathfrak{a} \in \mathcal{A}$
16:             $queue.add((s', \mathfrak{a}(\delta')), \mathbf{h}(\alpha(s, \mathcal{G}), \mathfrak{a}(\delta)))$
17:          **end for**
18:       **end if**
19:    **end while**
20: **end while**

---

indicates is the total resource allocated to finding a solution, such as time limit or the total number of iterations.

The algorithm begins by creating a priority queue and putting abstract edges from the initial state into the queue with their heuristic values as their priorities. At each iteration, the algorithm selects the abstract edge with the lowest heuristic value, and attempts to construct a successor state by sampling feasible continuous parameters for the abstract action in the associated state using the function SMPLCONT, a subprocedure in SAHS.

Because calling a motion planner is the most expensive computation in generating the successor state, the SmplCont function tries to reduce the number of motion planning calls. It first attempts $N_{smpl}$ times to sample $N_{mp}$ continuous parameter samples that satisfy constraints other than the existence of collision-free motion. For instance, it checks the existence of an IK solution and for collisions at pick or place base poses for a pick-and-place action. We call the samples that satisfy these cheap-to-evaluate constraints *partially feasible samples*. If we cannot find any partially feasible samples, then SmplCont returns an empty set. Otherwise, for each sample in the partially feasible samples, we call the motion planner to see if there exists a feasible motion. If there is, then we return that value.

If SMPLCONT returns feasible continuous parameters, we simulate the successor state using the transition model. If the successor state is in the goal set, it returns the plan by retracing the plan to the root. Otherwise, it checks whether the plan length is less than the planning horizon, and if so, adds the abstract edges to the queue. If it fails to sample feasible continuous parameters, it moves onto the next abstract edge on the queue. Unlike discrete graph search, our search space involves continuous values, so when the queue is empty we add the initial abstract edges back to the queue so that we can revisit abstract action sequences, while sampling different continuous parameter values.

This algorithm, as given, is not probabilistically complete. To guarantee probabilistic completeness, we need to revisit each sequence of abstract actions infinitely often with increasing effort to sample continuous parameters. We can do this with a slight modification to SAHS, but we present this particular version which is more intuitive to understand. The probabilistically complete version of our algorithm, along with its proof, is given in the appendix.

## 5. Learning to rank abstract actions

The heuristic function shown in (1) prioritizes abstract edges primarily based on the number of objects to move in a state. Its prioritization of abstract actions *within* a state, however, is rather nave: it only discourages the given abstract action if it tries to move a goal object already in its goal region.

We may, using our abstract representation, hand-design a method for prioritizing abstract actions, such as preferring an abstract action that manipulates a reachable object or an object that occludes many other objects. However, we found this to be ineffective and tedious.

Alternatively, we could learn an action-value function, and then use the action-values to prioritize abstract actions choices. However, this is data inefficient because to obtain accurate values of all of the actions of a state, we would need the target values for many of them. However, in our training data, we only have values of actions that led to the goal. Thus, we take a different approach.

Rather than estimating action values, we use a ranking function that deliberately discourages taking actions missing from the training data. More concretely, a ranking function takes as inputs a pair of abstract problem representation and actions, and outputs a *rank value* for that pair. The rank among the abstract actions within the abstract state is determined by this rank value: the higher the rank value, higher the ranking.

Our learning algorithm assumes that we have plans that were obtained by solving past problem instances, where each plan is a sequence of state–action pairs,

$$[(s_0, \mathfrak{a}_0(\delta_0, \kappa_0)), \ldots, (s_{T-1}, \mathfrak{a}_{T-1}(\delta_{T-1}, \kappa_{T-1})), (s_T, \emptyset)]$$

Here, $s_T \in \mathcal{G}$ and $\mathcal{G}$ denotes the goal for which that plan was made. From this, we can construct $T + 1$ supervised training examples of the form $(\alpha(s_t, \mathcal{G}), \mathfrak{a}_t(\delta_t))$. Aggregating this data from multiple start–goal pairs, and partitioning it according to each abstract action type $\mathfrak{a}$, we end up with a dataset $\mathcal{D}_\mathfrak{a}$ for each $\mathfrak{a}$, with entries of the form $(\alpha(s_t, \mathcal{G}), \delta)$.

Denote the rank function for action type $\mathfrak{a}$ parameterized by $\theta$ as $\hat{F}_\mathfrak{a}(\,\cdot\,,\,\cdot\,; \theta)$. We wish to rank the abstract actions such that the actions that appeared in past plans have higher rank values. This is implemented using the following large-margin loss

$$\mathcal{L}_{LM}(\theta) = \sum_{(s,\mathcal{G},\delta) \in \mathcal{D}_a} \max\left(0, 1 - M(\alpha(s,\mathcal{G}),\delta;\theta)\right)$$

where

$$\begin{aligned}M(\alpha(s,\mathcal{G}),\delta;\theta)\\= \widehat{F}_a(\alpha(s,\mathcal{G}),\delta;\theta) - \max_{\delta' \in \Delta \setminus \{\delta\}} \widehat{F}_a(\alpha(s,\mathcal{G}),\delta';\theta)\end{aligned}$$

Let us examine this objective function for a single data tuple $(s,\mathcal{G},\delta)$. Intuitively, the function is penalizing $\widehat{F}_a$ when the difference between the rank value of $\delta$ and the max value of the rest of the actions, $\Delta \setminus \{\delta\}$, is smaller than 1. The term $\max\left(0, 1 - M(\alpha(s,\mathcal{G}),\delta;\theta)\right)$ grows linearly with respect to this difference if the difference is smaller than 1. Otherwise, it evaluates to 0.

## 5.1. Augmenting the hand-designed heuristic function

As mentioned, the heuristic function H in (1) is primarily evaluating states but not the actions within a state. We now show how we can use our rank function to evaluate the actions within a state. One naïve way is to simply add the rank value, $\widehat{F}_a(\alpha(s,\mathcal{G}),\delta;\theta)$ to H. However, the rank value can be any real number, and may override our heuristic function. We wish to avoid this because our rank function is trained by comparing values of actions within a state, not across states. Thus, we scale the rank values so that they take a value between 0 and 1. The abstract-edge heuristic function that we use is

$$\begin{aligned}\mathbf{H}_{rank}(\alpha(s,\mathcal{G}),\mathfrak{a}(\delta)) = \mathrm{H}(\alpha(s,\mathcal{G}))\\- \frac{\exp\left(\widehat{F}_a(\alpha(s,\mathcal{G}),\delta;\theta)\right)}{\exp\left(\sum_{\alpha',\delta'} \widehat{F}_{a'}(\alpha(s,\mathcal{G}),\delta;\theta)\right)}\end{aligned} \quad (2)$$

Let us examine this heuristic function. The first term is our hand-designed heuristic function, as defined in (1). The second term is responsible for ranking abstract actions within a state, normalized to be between 0 and 1; the higher the rank value, the lower the heuristic value for that action.

To use this with SAHS, we put abstract edges in the search queue according to $\mathbf{H}_{rank}$ values. For abstract edges with the same state, an abstract edge that has the highest $\widehat{F}_a(\alpha(s,\mathcal{G}),\delta;\theta)$ value will be explored first. For abstract edges with different states, the abstract edge with a state that has the highest $\mathrm{H}(\alpha(s,\mathcal{G}))$ value will be explored first, and for all the abstract edges with the same state, the one with the with the highest $\widehat{F}_a(\alpha(s,\mathcal{G}),\delta;\theta)$ value will be explored first.

## 5.2. Representing the rank function

The rank function takes an abstract state representation as an input. The challenge in using the abstract representation $\alpha(s,\mathcal{G})$ with a NN is representing the abstract state in a form that can be used as an input to a NN and designing a NN architecture that effectively uses that representation. Because our abstract state is *relational*, one natural way to represent it is using a graph, where each node is associated with an entity, and each edge encodes the relationship among multiple entities. Typical NN architectures, such as fully connected NNs or convolutional NNs, cannot handle such graph and require a fixed-size vector as an input. Moreover, we need to collectively consider the relationships among *all* the entities, which requires information propagation from other entities to compute a value of a single entity.

We use GNNs to resolve these issues. GNNs take as an input a graph, computes messages at nodes and edges, performs rounds of message passing, and outputs quantities of interest. For us, the input to a GNN is a graph representation of an abstract state-and-goal, and the output is a set of rank values of moving object $o$ into region $r$ for all objects and regions.

One of the biggest design choices in using a GNN is the input graph structure. It must be designed such that it encodes all the necessary information, and the information flows to appropriate nodes. One distinctive aspect in our application is the ternary predicate, OCCLUDESMANIP, which takes three entities as an input. Encoding this requires a hypergraph, where each edge connects three nodes associated with the corresponding three entities.

Using our domain knowledge, we ensure that each edge is connected to nodes associated with the entities with correct types for the predicates on that edge. For instance, it is unnecessary to have an edge that connects three object nodes, because OCCLUDESMANIP requires two objects and a region. Thus, we construct a graph such that we have a separate component for each region, where for each component, we have $|\mathbf{O}^{(M)}| + 1$ number of vertices for all the objects and the associated region. Each edge connects two objects and a region, and encodes binary and ternary predicates. This representation allows us to define the edges in a hypergraph with a tensor of dimensions $|\mathbf{O}^{(M)}| \times |\mathbf{R}| \times d_e$, where $d_e$ is the dimensionality of an edge feature. An example of our graph is shown in Figure 6.

We now show how Node and Edge vectors in Figure 6 are defined. Define $x_{e_i}$ as a vector of unary predicate values for each entity $e_i \in \mathbf{O}^{(M)} \cup \mathbf{R}$,

$$x_{e_i} = [\text{ISOBJECT}(e_i), \text{ISREGION}(e_i), \text{ISGOAL}(e_i), \text{PREFREE}(e_i)]$$

For all ordered pairs of entities $e_i, e_j$, we define $x_{e_i e_j}$ as a vector of binary predicate values,

$$\begin{aligned}x_{e_i e_j} =\\\big[\text{INREGION}(e_i, e_j), \text{OCCLUDESMANIP}(e_i, e_j), \text{MANIPFREE}(e_i, e_j)\big]\end{aligned}$$

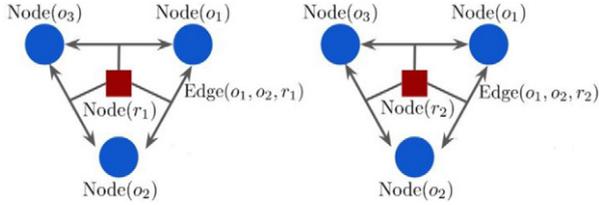

**Fig. 6.** The graph encoding the geometric predicates of a scene with three objects $o_1$, $o_2$, and $o_3$ and two regions $r_1$ and $r_2$. We represent a scene with a graph with two components, where each component of the graph is a fully connected graph associated with a region. On the left is a component associated with region $r_1$ and right is the one with associated with $r_2$. Each Edge($o_i$, $o_2$, $r_k$) encodes the set of binary predicates associated between two objects $o_i$ and $o_j$, such as INREGION($o_i$, $o_j$), and the ternary predicate associated with the two objects and region $r_k$, OCCLUDESMANIP($o_i$, $o_j$, $r_k$).

For all entities $e_i$, $e_j$ and region $r_k$, we define $x_{e_i e_j r_k}$ as a ternary relation value

$$x_{e_i e_j r_k} = [\text{OCCLUDESMANIP}(e_i, e_j, r_k)]$$

We define Node($e_i$), as

$$\text{Node}(e_i) = x_{e_i}$$

At an edge between objects $o_i$ and $o_j$ on the component associated with region $r_k$, we have an edge vector Edge($o_i$, $o_j$, $r_k$) defined as

$$\text{Edge}(o_i, o_j, r_k)$$
$$= [x_{o_i}, x_{o_j}, x_{o_i o_j}, x_{o_j o_i}, x_{o_i r_k}, x_{o_j r_k}, x_{o_i o_j r_k}, x_{o_j o_i r_k}]$$

Within each component of the graph shown in Figure 6, we see that the graph is fully connected. The reason for this is to minimize the inference time. For the GNN with a given number of edges and nodes, this amounts to minimizing the number of message passing rounds while propagating all the necessary information. With this fully connected structure within each component, two rounds of message passing is sufficient to propagate the necessary messages.

To see this, consider an abstract action that moves object $o$ to region $r$. To evaluate this action, we need to consider the following three factors based on its direct neighbors. (1) Is $o$ already in $r$? (2) What objects or regions does $o$ occlude? (3) What objects is $o$ occluded by? If we make the graph fully connected, all of these factors can be considered with a single message-passing round. Equally important for evaluating moving $o$ to $r$ is the information from indirect neighbors: is the object that occludes $o$ or occluded by $o$ in turn occluding or occluded by some other object(s) $o'$? For our graph structure, two rounds of message passing is sufficient to propagate these pieces of information.

The details of the computations in our GNN, such as what aggregator function we use and how we define the node and edge embedding functions, are included in the appendix.

# 6. Learning biased samplers for continuous parameters

So far, we have described our method for guiding the search for abstract actions. We now describe our strategy for learning a sampler to guide the search for continuous parameters. We begin with our relaxed problem representation based on key configurations, which is used as an input to our sampler.

## 6.1. Problem representation using key configurations

To improve planning efficiency, the continuous parameters predicted by a learned sampler should satisfy the feasibility constraints. Typically, these continuous parameters implicitly represent the goal configuration for the low-level motion planner, such as a base pose and grasp for picking an object. Thus, to generate promising values, we need a representation of the scene that enables a NN to infer the existence of a collision-free motion.

Ideally, we would use C-space obstacles, but this generally is prohibitively expensive to compute. Thus, we approximate the collision information at essential regions of the C-space, using a set of key configurations. Key configurations are a set of configurations that we construct by first collecting from our planning experience a set of robot configurations that were used in the plan solutions, and then subsampling them using a threshold on the distances among them. Given a scene, we represent the state using the collision information at these key configurations. Our insight is that we do not need to completely construct the C-space obstacles, but only at essential regions of the C-space that the robot is likely to re-use in the future problem instances.

Another essential piece of information that we need to capture is the goal for our original problem. We use binary vector representation of $V_{\text{manip}}(\mathfrak{o}(o_G, r_G))$ for all goal object–region pairs mentioned in the $\mathcal{G}$, $(o_G, r_G)$, to encode this. For each configuration in $V_{\text{manip}}(\mathfrak{o}(o_G, r_G))$, we check whether the distance to the closest key configuration is below a threshold, and set it to 1 and otherwise to 0. While this may not be the actual swept-volume for moving the goal object to the goal region, it represents an approximation of it.

Combining these two types of information, our key-configuration-based problem representation is an $n_k \times 2$ matrix whose first column is a binary vector representing the collision information, and the second column representing $V_{\text{manip}}(\mathfrak{o}(o_G, r_G)) \forall o_G, r_G \in \mathcal{G}$. Figure 7 shows an example of state and goal representation based on a subset of key configurations.

## 6.2. Learning a biased sampler from planning experience

Now that we have a representation, we describe the learning algorithm for the sampler. We are given the same sequence

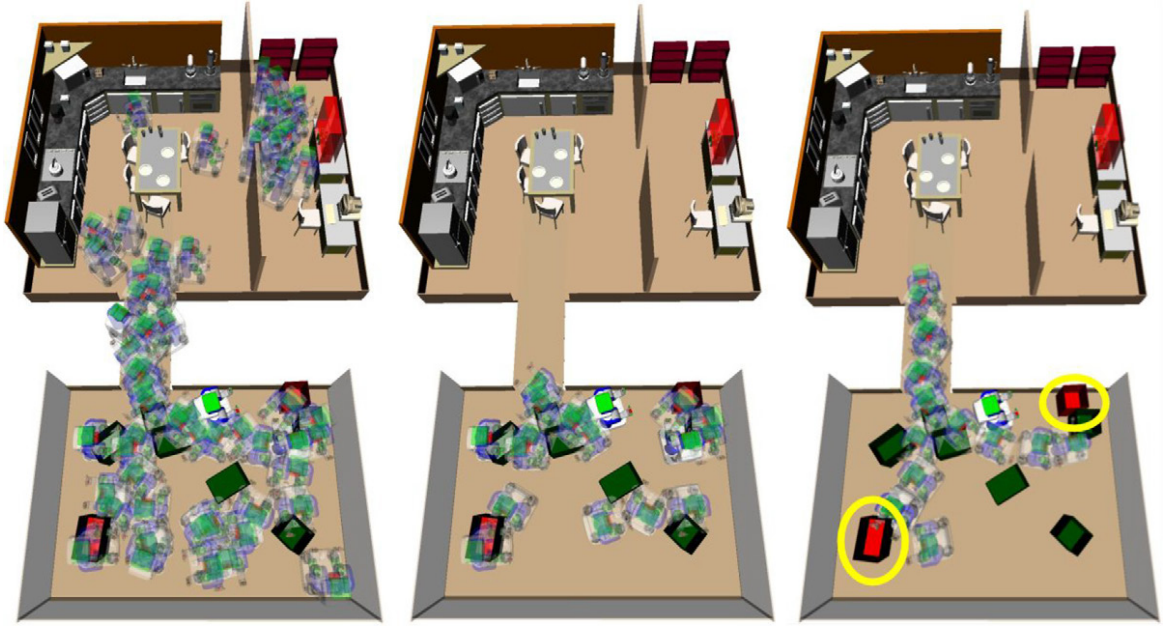

**Fig. 7.** (Left) A subset of key configurations used for this domain. (Middle) Key configurations that are in collision. (Right) Key-configuration-representation of $V_{\text{manip}}(\mathfrak{o}(o_G, r_G))$ for the two goal red objects marked yellow circles. The state and goal are represented as a binary vector of shape $n_k$ by 2, where the first column encodes collision information shown in the middle figure, and the second column encodes $V_{\text{manip}}(\mathfrak{o}(o_G, r_G))$ shown in the right-most figure.

of state–action pairs that we considered in the previous section,

$$[(s_0, \mathfrak{a}_0(\delta_0, \kappa_0)), \ldots, (s_{T-1}, \mathfrak{a}_{T-1}(\delta_{T-1}, \kappa_{T-1})), (s_T, \cdot)]$$

in which $s_T \in \mathcal{G}$. Like the large-margin objective in (2), we use the pessimism principle to discourage actions missing from data. Unlike (2), however, we cannot minimize the maximum value of the missing actions, because we have a continuous space for parameters.

One observation we can make is that we do not have to assign low values to *all* of the continuous parameters that are not in the dataset, but only to those that our sampler generates. Thus, we assign arbitrary low values to the continuous parameters if they are being generated by our sampler and are very different from those in our dataset. We implement this intuition using GANs.

In a GAN, we train a *discriminator*, which evaluates state and continuous parameter pairs based on how frequently they appeared in the data. We then train a *generator*, which in our case is the sampler, using the discriminator as its objective function. The objective function for the discriminator, $D$, is

$$\min_D \mathbb{E}_{s \sim P_S} \left[ \mathbb{E}_{\kappa \sim P_{K|S=s}}[\log D(\phi(s)), \kappa)] \right] + \mathbb{E}_{\kappa \sim P_\theta}[\log (1 - D(\phi(s), a))]$$

where $\phi(s)$ represents the state and goal represented with key configurations, $P_{K|S=s}$ is the desired distribution over continuous parameters in the given state, and $P_\theta$ is our learned distribution over continuous parameters given $s$.

Let us examine this objective function. The first term is trying to increase the values of state-and-action pairs in our data, which is similar to (2); now, if we are to follow the same intuition as (2), we would have to have a term that decreases the maximum value of the rest of the actions.

Unfortunately, we have a continuous space, and this is difficult to enforce. Thus, instead, the second term in this objective function minimizes the values of parameters generated by the learned sampler. This allows assigning arbitrarily low values to the generated actions if they are very different from those in the data. When they are similar, the first term prevents it from assigning arbitrarily low values. Using $D$ as an objective function, we train our sampler

$$\max_\theta \mathbb{E}_{\kappa \sim P_\theta(K|s)}[D(\phi(s), \kappa)]$$

The two networks are trained in an alternating fashion, and under a specific set of assumptions, we can show that $P_\theta$ converges to $P_{K|S}$ (Goodfellow et al., 2014).

In practice, however, the adversarial nature of this training scheme makes training unstable. Thus, unlike our previous work, we use Wasserstein GANs (WGANs) (Arjovsky et al., 2017), which have proven to have more stable training behavior than GANs. Our objective function for training the distribution $P_\theta$ that approximates $P_{K|S}$ is

$$\max_{D \in \|D\|_L \leqslant 1} \mathbb{E}_{s \sim P_S} \left[ \mathbb{E}_{\kappa \sim P_{K|S=s}}[D(s, \kappa)] - \mathbb{E}_{\kappa \sim P_\theta}[D(s, \kappa)] \right]$$

where $\| D \|_L \leqslant 1$ indicates the set of all 1-Lipschitz functions $D : \mathcal{S} \times K \to \mathbb{R}$.

**Algorithm 3** WGAN-GP($\mathbf{D}_K, \lambda, n_{tot}, n_c, n_b, lr_\theta, lr_\alpha$)

---

**for** $t = 0$ **to** $n_{tot}$
  **for** $t_c = 0$ **to** $n_c$
    **for** $t_m = 0$ **to** $n_b$
      $(s^{(i)}, \kappa^{(i)}) \sim \mathbf{D}_K$
      $z^{(i)} \sim P_Z(z), \kappa_\theta^{(i)} \sim \pi_\theta(s^{(i)}, z^{(i)})$
      $\epsilon^{(i)} \sim U[0,1], \hat{\kappa}^{(i)} = \epsilon \kappa^{(i)} + (1 - \epsilon) \hat{\kappa}^{(i)}$
      $L^{(i)} = D_\alpha(s^{(i)}, \kappa^{(i)}) - D_\alpha(s^{(i)}, \kappa_\theta^{(i)}) +$
      $\lambda \left( \|\nabla_{\hat{\kappa}^{(i)}} D(s^{(i)}, \hat{\kappa}^{(i)})\|_2 - 1 \right)^2$
    **end for**
    $\alpha = \alpha + \text{Adam}(lr_\alpha, \nabla_\alpha \frac{1}{n_b} \sum_{i=1}^{n_b} L^{(i)})$
  **end for**
  $\{z^{(i)}\}_{i=1}^{n_b} \sim P_Z(z)$
  $\theta = \theta + \text{Adam}(lr_\theta, \nabla_\theta \frac{1}{n_b} \sum_{i=1}^{n_b} f(s^{(i)}, \pi(s^{(i)}, z^{(i)})))$
**end for**
**return** $\pi_\theta$

---

We represent the function class $\|D\|_L \leqslant 1$ with NNs parameterized by $\alpha$, which we denote by $D_\alpha$. To enforce the 1-Lipschitz constraint on these NNs, the original WGAN used hard gradient clipping which leads to difficulty in optimization.

Instead, we use WGAN with gradient penalty (WGAN-GP) (Gulrajani et al., 2017). This method uses a soft constraint on the norm of the gradient of the function $D_\alpha$ based on the observation that a differential function is 1-Lipschitz if and only if its gradients have norms of at most 1 everywhere. Our objective for training a discriminator is

$$\min_\alpha \mathbb{E}_{s \sim P_S} [\mathbb{E}_{\kappa \sim P_{K|S=s}} [D_\alpha(s, \kappa)] + \mathbb{E}_{\kappa \sim P_\theta(K|s)} [D_\alpha(s, \kappa)]$$
$$+ \lambda \cdot \mathbb{E}_{\hat{\kappa} \sim P_{\hat{K}}} \left[ (\| \nabla_{\hat{\kappa}} D_\alpha(s, \hat{\kappa})\|_2 - 1)^2 \right] ]$$

Here, the last term is responsible for softly enforcing the Lipschitz constraint. Since enforcing the constraint everywhere is intractable, WGAN-GP enforces it only on the samples from $P_{\hat{K}}$, where $P_{\hat{K}}$ is defined as a uniform distribution on a straight line between a pair of samples from $P_{K|S}$ and $P_\theta$.

The pseudo-code of our learning algorithm is shown in Algorithm 3. It takes in as inputs the training dataset $\mathbf{D}_K$, gradient penalty scale term $\lambda$, total number of iterations, $n_{tot}$, number of gradient steps for discriminator training at each iteration, $n_c$, the batch size $n_b$, and learning rates for the sampler and discriminator, $lr_\theta$ and $lr_\alpha$, respectively. It first begins by training the discriminator. At each iteration of discriminator training, it creates a batch of $\hat{\kappa}$ values, by sampling a point from $\mathbf{D}_K$, and generating a point from $\pi_\theta$. It then samples a random number uniformly between 0 and 1 and uses this as a weight to mix the point from $\mathbf{D}_K$ and point from $\pi_\theta$. These are used to compute our objective function for each point in our batch, $L^{(i)}$. Once all of these values are computed, then we take a gradient step using the Adam optimizer (Kingma and Ba, 2015). We repeat these steps $n_c$ times and then update sampler. The entire process is repeated $n_{tot}$ number of times.

One of the fundamental challenges in learning a generative model is evaluating a trained model. For its typical application of generating images, measuring the quality of a generated image is difficult. Fortunately for us, the action space is relatively low dimensional, and we can use kernel density estimation (KDE), which is too expensive to use during inference time, but fast enough for evaluation, to evaluate the quality of the trained model. To do this, for each state in our dataset, we generate 100 samples and then fit it with KDE. We then measure the likelihood of the continuous parameters for that state using KDE. We average the likelihood across all states in the dataset, and discard the trained weights if the average likelihood value are too low. We also perform data cleaning because our data is generated by a sampling-based algorithm. The details are in the appendix.

## 7. Experiments

### 7.1. Domain description

We consider two different environments. One is the box-moving domain shown in Figure 1(left) where the objective is to move a set of boxes from their initial locations to the kitchen region. The second environment is the cupboard domain where the robot has to move a target object from the cupboard to the packing box shown in Figure 1(right). A problem instance consists of an initial state, defined by the poses of movable objects, and the goal represented with a conjunction of INREGION predicate instances. Problem instances are generated by randomly sampling the initial object poses and randomly choosing goal objects.

The numbers of movable objects are 8 and 10 for box-moving and cupboard domains, respectively. Thus, the state-space for the box-moving domain is $\text{SE}(2)^8 \times \mathcal{C}_{\text{base}} \times \mathcal{C}_{\text{left−arm}} \times \mathcal{C}_{\text{right−arm}}$ for the combined configurations of eight boxes, and the configurations of two arms and base. For the cupboard domain, it is $\text{SE}(2)^{10} \times \mathcal{C}_{\text{base}} \times \mathcal{C}_{\text{right−arm}}$ for the combined configurations of 10 small cupboard objects, and the configurations of a single arm and base.

To make sure that each problem instance is non-trivial (i.e., cannot be solved by simply moving just the goal objects to goal regions), we define the distribution over initial object poses such that the robot must manipulate at least two or three objects in addition to the goal objects. In the box-moving domain, we do this by randomly placing at least three or four objects at the exit and around the robot. Similarly, in the cupboard domain, we randomly place the goal object at the back of the cupboard to ensure the robot must rearrange at least two or three objects.

Although our framework can be applied to cases where we have multiple manipulation operators, we focus on the pick-and-place operator in our experiment. For the box-moving environment, we have two-arm pick-and-place; for the cupboard environment we have one-arm pick-and-place. The pick operation is defined by two continuous parameter vectors: the first is a grasp vector, denoted with $(d, h, \chi)$, that specifies a depth $d$, as a fraction of the size of the object in the approach direction, height $h$, as a a fraction of object height, and angle in the approach direction $\chi$, respectively. Here $d$ and $z$ have a range $[0.5, 1]$, and $\chi$ has a range $[\frac{\pi}{4}, \pi]$. The second is the pick base pose vector, $(x_r^o, y_r^o, \psi_r^o)$, which represents the robot base pose relative to the pose of an object $o$, whose pose in global frame is $(x_o, y_o, \psi_o)$. For the two-arm case, the grasp parameters $(d, h, \chi)$ specify the mid-point of the two grippers and for the one-arm case they specify the mid-point of the right arm's gripper. The continuous parameters of place operation is specified by the robot base pose in the world reference frame, and assume that the arms are fixed when the robot's base is moving.

Given an object, a grasp and base pose for pick, an IK solver is used to generate collision-free arm configurations for picking at the specified base pose and grasp parameters. In the box-moving environment, once we determine the pick-and-place continuous parameters, we plan the base motion plans for both picking and placing the objects. In the cupboard environment, we omit motion planning and simply check collisions at the pick and place configurations, which is sufficient in this domain.

We use OpenRAVE simulator (Diankov, 2010) with the Flexible Collision Checking library. We did not simulate physics. We use bidirectional rapidly-exploring randomized trees (biRRT) as our motion planner (Kuffner and LaValle, 2000), and FastIK as our IK solver. We use 1,000 planning episodes for training samplers in the cupboard domain, and 1,500 planning episodes for training samplers in the box-moving domain. For training the ranking function, we use 250 planning episodes. The number of key configurations is 355 for the cupboard domain, and 618 for the box-moving domain.

## 7.2. Benchmark description

The claims in this paper are summarized as follows.

- *Claim 1 (computational efficiency and robustness)*: learning to guide planning is more computationally efficient than pure planning and more robust than pure learning.
- *Claim 2 (data efficiency)*: for learning from planning experience, using pessimism against the actions missing from data is more data efficient than estimating the cost-to-go.
- *Claim 3 (effective problem representation design)*: It is more effective to use a relaxed problem representation as an input to the search guidance predictors than the full problem representation.
- *Claim 4 (generalization capability of our representation)*: for the task guidance, we can generalize across different environments while for continuous parameters we can generalize within an environment. For both cases, they generalize to different goals; in particular, we can train our predictors from easier instances of the problem and generalize to more difficult problems that involve longer planning horizon.

To support *Claim 1 (computational efficiency and robustness)*, we compare against the following benchmarks.

- RANK-WGANGP: A pure-learning method that uses the abstract action that has the highest rank value with respect to $\widehat{F}$ and the continuous parameters predicted by a sampler learned using WGAN-GP without planning. It resets to the initial state if it samples an infeasible action.
- IRSC: Iterative Resolve Spatial Constraint. It is a pure-planning algorithm. We extend RSC (Stilman et al., 2007), which is the state-of-the-art algorithm for manipulation among movable obstacles, to moving multiple objects to a goal region. To do this, we first plan pick-and-place motions for moving goal objects to goal regions with checking collisions only at object placements and initial and final robot configurations. From this, we get an order to pack objects into the goal regions. Each single-object packing subproblem is solved by an application of RSC. If, after some number of iterations, RSC does not find a solution, we modify the object ordering and try randomly permuting the unplaced boxes. The algorithm will eventually try all orderings if given enough time.
- SAHS-HCOUNT: A pure-planning algorithm based on SAHS that uses the hand-designed abstract-edge heuristic function shown in (1) and a uniform sampler for continuous parameters.

We compare these with our guided versions of SAHS.

- SAHS-RANK: uses the abstract-edge heuristic function in (2). Here $\widehat{F}$ is trained with our abstract problem representation and the large-margin loss. Uses a uniform sampler for continuous parameters.
- SAHS-RANK-WGANGP: same as SAHS-RANK but uses a sampler trained using key-configuration problem representation and WGAN-GP objective function.

To support *Claim 2 (data efficiency)*, we have the following benchmarks that estimate the cost-to-go of a given state-and-action instead of a pessimism based loss:

- SAHS-MSE: uses the abstract-edge heuristic function in (2). Here $\widehat{F}$ is trained with our abstract problem representation but mean-squared-error loss with the remaining number of steps to the goal as target values. Uses a uniform sampler for continuous parameters.
- SAHS-RANK-ACTORCRITIC: same as SAHS-RANK but uses a sampler trained with key-configuration problem representation and a variant of ddpg (Lillicrap et al., 2016). The critic is trained with the planning experience dataset, with remaining number of steps to the goal as target values. The sampler is trained by optimizing this critic.

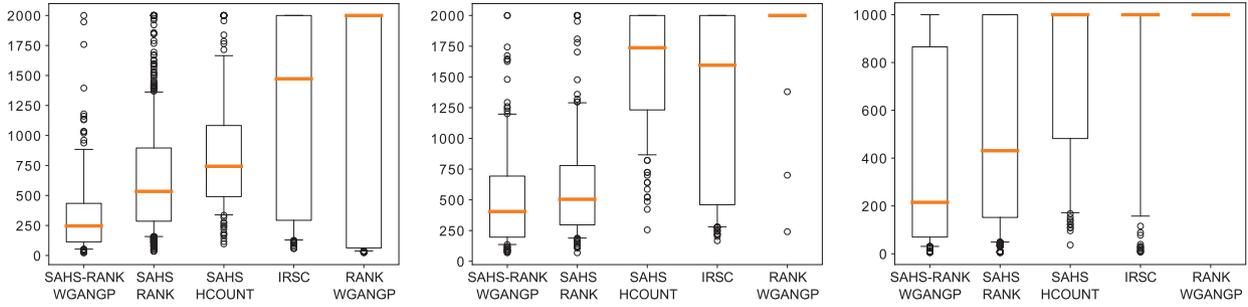

**Fig. 8.** Box plots for planning times for (left) moving one box in the box-moving domain, (middle) for moving four boxes in the same domain, and (right) for moving a target object in the cupboard domain. The box-moving domain had 2,000 seconds time limit, and the cupboard domain has 1,000 seconds time limit. Boxes indicate 25th and 75th percentiles, and Whiskers indicate 10th and 90th percentiles.

**Table 1.** Success rates of different algorithms in the box-moving domain with 2,000 s time limit (left) and cupboard domain 1,000 s time limit (right)

| Algorithm | Success rate (1 box, 2,000 s) | Success rate (4 boxes, 2,000 s) | Algorithm | Success rate (1 obj, 1,000 s) |
|---|---|---|---|---|
| IRSC | 0.53 | 0.51 | IRSC | 0.21 |
| RANK-WGANGP | 0.43 | 0.03 | RANK-WGANGP | 0.00 |
| SAHS-HCOUNT | 0.94 | 0.56 | SAHS-HCOUNT | 0.44 |
| SAHS-RANK | 0.94 | 0.96 | SAHS-RANK | 0.75 |
| SAHS-RANK-WGANGP | **0.99** | **0.97** | SAHS-RANK-WGANGP | **0.81** |

**Table 2.** Various versions of guided-SAHS. Those marked with a dash do not use the learned sampler or $\widehat{F}$

| Algorithms | $\widehat{F}$ loss | $\widehat{F}$ representation | Sampler Loss | Sampler representation |
|---|---|---|---|---|
| SAHS-HCOUNT | — | — | — | — |
| SAHS-RANK | Large-margin | Abstract | — | — |
| SAHS-RANK-WGANGP | Large-margin | Abstract | WGAN-GP | Key configurations |
| SAHS-MSE | MSE | Abstract | — | — |
| SAHS-RANK-ACTORCRITIC | Large-margin | Abstract | Variant of DDPG | Key configurations |
| SAHS-POSERANK | Large-margin | Full problem | — | — |
| SAHS-RANK-POSEWGANGP | Large-margin | Abstract | WGAN-GP | Full problem |

To support *Claim 3 (effective problem representation design)*, we compare against the following benchmarks that use the full problem representation based on poses and shapes of movable objects, instead of a relaxed problem representation.

- SAHS-POSERANK: uses the abstract-edge heuristic function (2). Here $\widehat{F}$ is trained with the full problem representation and large-margin loss. Uses a uniform sampler for continuous parameters.
- SAHS-RANK-POSEWGANGP: same as SAHS-RANK but uses a sampler trained with full problem representation and WGAN-GP objective function.

Table 2 summarizes the various versions of guided SAHS.

To support *Claim 4 (generalization capability of our representation)*, we only collect planning experience from the box-moving domain, where the goal is to move a single box to the kitchen. Then, we test its performance in two domains without retraining. First, in the same box-moving domain, but where the goal is changed to moving four boxes to the kitchen, instead of one. This demonstrate its capability to generalize to harder problems. Second in the cupboard domain, where the goal is to move smaller objects into a packing box using one-arm pick-and-place, to demonstrate its capability to generalize across environments with different geometric details.

To evaluate each of these algorithms, we measure different quantities. For comparing pure planning and pure learning methods with guided planners, we measure the planning time to find a solution within a time limit. For comparing different instantiations of SAHS that uses different abstract-action heuristic function and continuous parameter samplers, we measure the number of nodes to find a solution within a limit on the total number of explored nodes. In both cases, we measure the success rate defined by the percentage of problem instances solved within the given resource limit. To measure these quantities, we test the algorithms on 25 problem instances in each setup. For algorithms that involve planning, we use five different planning seeds. For algorithms that involve learning, we additionally use five different training seeds. The details of hyperparameters and training data used for these benchmarks are described in the appendix.

## 7.3. Results

### 7.3.1. Claim 1 (computational efficiency and robustness).

Figure 8(left) shows the planning time results for moving a single box in the box-moving domain. We see that the median of SAHS-RANK is about 3 times faster than that of IRSC and about 1.5 times faster than that of SAHS-HCOUNT. Further, the median of SAHS-RANK-WGANGP is 6 times faster than IRSC and 3 times faster than SAHS-HCOUNT. The guidance-based approaches have much lower 90th percentiles as well, indicating that they are especially better for harder problem instances. IRSC performs badly because it makes the *monotonicity* assumption, which states that problems can be solved by touching each object only once, and this does not hold in this problem. RANK-WGANGP performs the worst among all the methods, owing to its inability to overcome prediction mistakes. This also is evident in Table 1(left, second column). Although the guidance-based approaches, including SAHS-HCOUNT which is guided by a hand-designed function, solve more than 90% of the problems, pure-learning (RANK-WGANGP) and pure-planning (IRSC) solve only about half of the problems compared to the guidance-based approaches.

Figure 8(middle) shows the results for a harder problem with longer horizon where the robot has to pack four boxes. The median of SAHS-RANK-WGANGP is about 5.8 times faster than those of SAHS-HCOUNT and IRSC, and SAHS-RANK is about 3.5 times faster. The 90th percentile is again significantly lower than the benchmarks. It is also worth noting the significant drop in the success rate of the pure-learning approach in Table 1(left, third column). The main reason is that, because this is a longer-horizon problem than the one-box-moving scenario, there is more room for making prediction mistakes. In terms of success rates, the guidance based algorithms outperform the benchmarks.

Figure 8(right) shows the results in another problem for a cupboard domain where the robot has to rearrange small objects inside a shelf and then pack a target object to the green box. We again see that the guidance-based approaches are superior to pure planning and pure learning in terms of the median of planning times. IRSC especially suffers in this domain because the environment is tighter than the box-moving domain, making more instances of non-monotonic problems. Similarly, the tightness in the environment requires longer horizon plans, making rank-wgangp suffer. The success rates in Table 1(right) indicates that guidance-based algorithms, SAHS-RANK-WGANGP and SAHS-RANK, significantly outperform the benchmarks and improves the success rate by a factor of 1.8.

### 7.3.2. Claim 4 (generalization capability of our representation).

This claim is more precisely divided into three points: (1) abstract state-and-goal representation can generalize to different environments, (2) both abstract and key-configuration-based representations can generalize from easier problem instances to harder instances with longer planning horizon, and (3) both of them can generalize to variations within an environment.

To evaluate the first point, consider Figure 8(right), which shows results for a cupboard domain by applying the same ranking function that we learned from the box-moving domain without retraining. As noted previously, we see that the SAHS-RANK outperforms the benchmarks except SAHS-RANK-WGANGP, without retraining, indicating it can generalize to different environments.

To support the second point, consider Figure 8(middle) where we consider the box-moving domain with four goal-boxes. Here, we apply the sampler and rank function that we learned from problems with a single goal-box without retraining. As we have noted, SAHS-RANK and SAHS-RANK-WGANGP outperform the benchmarks without retraining, indicating they can generalize to harder problems.

The third point is already demonstrated since each problem is defined by different poses of objects within an environment.

### 7.3.3. Claim 2 (data efficiency).

Figure 9 shows the results of different losses for the single-goal-box problem in the box-moving domain. By comparing SAHS-RANK and SAHS-MSE, we can see that even with the same number of training data, we can solve problems twice as fast using ranking loss instead of MSE: the median number of nodes are 58 and 31, respectively. This indicates that the ranking loss is more data efficient than MSE, which does not penalize actions that are missing in the training data. This claim is further supported in Table 3, which shows that with the same amount of training data and limit on the number of nodes, SAHS-RANK 36% more problem instances than SAHS-MSE.

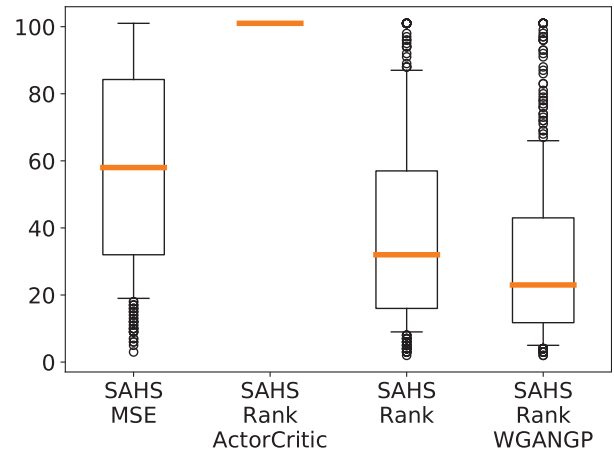

**Fig. 9.** Comparison of number of nodes to find a solution by various algorithms trained with different losses. The maximum number of nodes is set to 100.

**Table 3.** Success rates of SAHS guided by learning algorithms that use different losses and representations

| Algorithm | Success rate |
|---|---|
| SAHS-MSE | 0.55 |
| SAHS-RANK-ACTORCRITIC | 0.00 |
| SAHS-POSERANK | 0.63 |
| SAHS-RANK-POSEWGANGP | 0.50 |
| SAHS-RANK | 0.91 |
| SAHS-RANK-WGANGP | **0.97** |

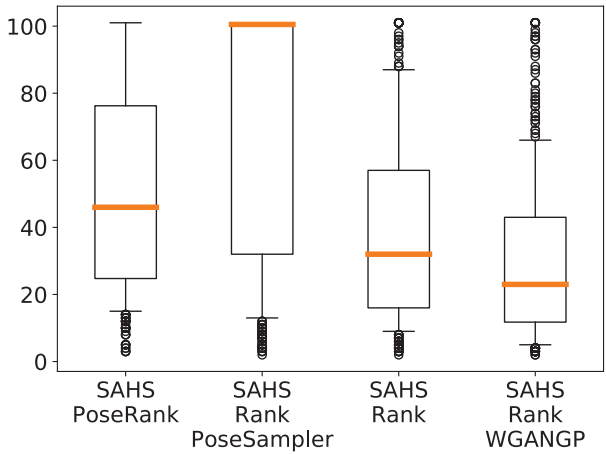

**Fig. 10.** Comparison of number of nodes to find a solution by various algorithms trained with different representations. The maximum number of nodes is set to 100.

We see a more pronounced difference for sampler losses. In Table 3, we see that SAHS-RANK-ACTORCRITIC could not solve any of the problems compared with SAHS-RANK-WGANGP, which solves 97% of the problems. This is because unlike an RL setting, we have a fixed dataset and the agent does not get to explore the environment by trying different actions. As a result, the values of actions missing in the data are extremely wrong, and sometimes are encouraged by the learned sampler. Most of the time, SAHS-RANK-ACTORCRITIC wastes its computations by trying infeasible continuous parameters.

*7.3.4. Claim 3 (effective problem representation design).* Figure 10 shows the results of different representations for the single-goal-box problem in the box-moving domain. For the task-level guidance, with the same amount of data, SAHS-POSERANK has 46 median number of explored nodes, compared with 32 of SAHS-RANK; Table 3 shows that SAHS-POSERANK solves only 63% of problem instances, whereas SAHS-RANK solves 91% of them. This indicates by using the relaxed abstract problem representation, we can solve more problems faster than using the full problem representation, even with the same number of training data and limit on the number of explored nodes .

Similarly, we can see the effect of key-configuration-based representation as well. SAHS-RANK-POSEWGANGP has a median number of nodes of 100, whereas SAHS-RANK-WGANGP has about 25; the success rate of SAHS-RANK-POSEWGANGP, indicated in Table 3, is 50%, whereas for SAHS-RANK-WGANGP, it is 97%.

## 8. Discussion

We have presented a framework that integrates a novel planning algorithm, SAHS, with learning algorithms for abstract action ranking function and continuous parameter sampler. Although the planner itself is a generic search algorithm, our framework automatically learns domain-specific search guidance knowledge by leveraging its planning experience. We introduced our design principles for designing the representation and loss function for training these search guidance functions: relaxed problem representation and pessimism against actions missing from data. We showed that by using these principles, we can be more data and computationally efficient and aggressively generalize to harder problems and different environments. We also showed that by learning to *guide* planning, we can be more robust than pure-learning and more computationally efficient than pure-planning.

There are many important aspects of G-TAMP we are not addressing in this paper. First, we are not addressing the substantial problem of planning and execution under sensor uncertainty. We assume observable states and deterministic transitions, mainly because the problem is already hard enough even under these simplifying assumptions. To lift these assumptions, we would have to model the environment with a partially observable Markov decision process (POMDP), and use belief-space TAMP algorithms (Garrett et al., 2020; Kaelbling and Lozano-Pérez, 2013). In this setup, our design principles can still be applied but we would need a strategy for computing the abstract and key-configuration representations from a history of observations and actions.

Another important aspect we are not considering is optimality. We focused on *satisficing problems*, where the goal is to find a feasible solution rather than an optimal one. We believe if we are given a cost function, then we can modify SAHS to optimize it by taking into account the costs of the actions taken so far in addition to the heuristic and rank functions. However, we did not try this because (1) even the satisficing problem itself is hard enough for G-TAMP problems, and (2) computing a reasonable objective function is non-trivial. For example, a cost function that measures the total energy required to move all the objects, or the plan execution time, is difficult to compute.

## Funding

We gratefully acknowledge the support from Samsung Electronics (grant number IO210106-08290-01) and the Institute of Information and Communications Technology Planning and Evaluation (IITP) grant funded by the Korean Government (MSIT) (grant number 2019-0-00075), Artificial Intelligence Graduate School Program (KAIST).

## ORCID iDs

Beomjoon Kim  https://orcid.org/0000-0002-8888-7253
Luke Shimanuki  https://orcid.org/0000-0002-4156-9944
Leslie Pack Kaelbling  https://orcid.org/0000-0001-6054-7145
Tomas Lozano-Perez  https://orcid.org/0000-0002-8657-2450

## References

Argall BD, Chernova S, Veloso M and Browning B (2009) A survey of robot learning from demonstration. In: *Robotics and Autonomous Systems*.

Arjovsky M, Chintala S and Bottou L (2017) Wasserstein generative adversarial networks. In: *International Conference on Machine Learning*.

Battaglia PW, Hamrick JB, Bapst V, et al. (2018) Relational inductive biases, deep learning, and graph networks. *arXiv preprint arXiv:1806.01261*.

Cambon S, Alami R and Gravot F (2009) A hybrid approach to intricate motion, manipulation, and task planning. *The International Journal of Robotics Research* 28(1): 104–126.

Chitnis R, Hadfield-Menell D, Gupta A, et al. (2016) Guided search for task and motion plans using learned heuristics. In: *IEEE International Conference on Robotics and Automation*.

Chitnis R, Kaelbling LP and Lozano-Pérez T (2019) Learning quickly to plan quickly using modular meta-learning. In: *IEEE International Conference on Robotics and Automation*.

Chitnis R, Silver T, Kim B, Kaelbling LP and Lozano-Perez T (2020) CAMPS: Learning context-specific abstractions for efficient planning in factored MDPs. In: *Conference on Robotic Learning*.

Diankov R (2010) *Automated Construction of Robotic Manipulation Programs*. PhD Thesis, CMU Robotics Institute.

Domshlak C, Karpas E and Markovitch S (2010) To max or not to max: Online learning for speeding up optimal planning. In: *AAAI Conference on Artificial Intelligence*.

Driess D, Ha J and Toussaint M (2020a) Deep visual reasoning: Learning to predict action sequences for task and motion planning from an initial scene image. In: *Robotics: Science and Systems*.

Driess D, Oguz O, Ha J and Toussaint M (2020b) Deep visual heuristics: Learning feasibility of mixed-integer programs for manipulation planning. In: *International Conference on Robotics and Automation*.

Fink M (2007) Online learning of search heuristics. In: *Artificial Intelligence and Statistics*.

Garrett CR, Chitnis R, Holladay R, et al. (2021) Integrated task and motion planning. *Annual Review of Control, Robotics, and Autonomous Systems* 4: 265–293.

Garrett CR, Kaelbling LP and Lozano-Pérez T (2016) Learning to rank for synthesizing planning heuristics. In: *International Joint Conference on Artificial Intelligence*.

Garrett CR, Kaelbling LP and Lozano-Pérez T (2017) Sample-based methods for factored task and motion planning. In: *Robotics: Science and Systems*.

Garrett CR, Lozano-Peréz T and Kaelbling LP (2018) FFRob: Leveraging symbolic planning for efficient task and motion planning. *The International Journal of Robotics Research* 37(1): 104–136.

Garrett CR, Paxton C, Lozano-Pérez T, Kaelbling LP and Fox D (2020) Online replanning in belief space for partially observable task and motion problems. In: *IEEE International Conference on Robotics and Automation*.

Gilmer J, Schoenholz SS, Riley PF, Vinyals O and Dahl GE (2017) Neural message passing for quantum chemistry. In: *International Conference on Machine Learning*.

Goodfellow I, Pouget-Abadie J, Mirza M, et al. (2014) Generative adversarial nets. In: *Advances in Neural Information Processing Systems*.

Gori M, Monfardini G and Scarselli F (2005) A new model for learning in graph domains. In: *IEEE International Joint Conference on Neural Networks*.

Gravot F, Cambon S and Alami R (2005) Asymov: A planner that deals with intricate symbolic and geometric problems. In: *International Symposium on Robotics Research*.

Gulrajani I, Ahmed F, Arjovsky M, Dumoulin V and Courville AC (2017) Improved training of Wasserstein GANs. In: *International Conference on Machine Learning*.

Hauser K (2014) The minimum constraint removal problem with three robotics applications. *The International Journal of Robotics Research* 33(1): 5–17.

Helmert M (2006) The fast downward planning system. *Journal of Artificial Intelligence Research* 26(1): 191–246.

Hoffman J and Nebel B (2001) The FF planning system: Fast plan generation through heuristic search. *Journal of Artificial Intelligence Research* 14(1): 253–302.

Kaelbling LP and Lozano-Pérez T (2011) Hierarchical task and motion planning in the now. In: *IEEE Conference on Robotics and Automation*.

Kaelbling LP and Lozano-Pérez T (2013) Integrated task and motion planning in belief space. *The International Journal of Robotics Research* 32(9–10): 1194–1227.

Kim B, Kaelbling LP and Lozano-Pérez T (2018) Guiding search in continuous state-action spaces by learning an action sampler from off-target search experience. In: *AAAI Conference on Artificial Intelligence*.

Kim B, Kaelbling LP and Lozano-Pérez T (2019a) Adversarial actor–critic method for task and motion planning problems using planning experience. In: *AAAI Conference on Artificial Intelligence*.

Kim B, Kaelbling LP and Lozano-Pérez T (2019b) Learning to guide task and motion planning using score-space representation. *The International Journal of Robotics Research* 38(7): 793–812.

Kim B and Shimanuki L (2019) Learning value functions with relational state representations for guiding task-and-motion planning. In: *Conference on Robot Learning*.

King JE, Cognetti M and Srinivasa SS (2016) Rearrangement planning using object-centric and robot-centric action spaces. In: *IEEE International Conference on Robotics and Automation*.

Kingma D and Ba J (2015) Adam: A method for stochastic optimization. In: *3rd International Conference on Learning Representations (ICLR 2015)*.

Kingma DP and Welling M (2014) Auto-encoding variational Bayes. In: *International Conference on Learning Representations*.

Kipf TN and Welling M (2017) Semi-supervised classification with graph convolutional networks. In: *International Conference on Learning Representations*.

Krontiris A and Bekris KE (2015) Dealing with difficult instances of object rearrangement. In: *Robotics: Science and Systems*.

Kuffner J and LaValle S (2000) RRT-connect: An efficient approach to single-query path planning. In: *International Conference on Robotics and Automation*.

Li C, Chang W, Cheng Y, Yang Y and Poczos B (2017) MMD GAN: Towards deeper understanding of moment matching network. In: *Neural Information Processing Systems*.

Lillicrap TP, Hunt JJ, Pritzel AP, et al. (2016) Continuous control with deep reinforcement learning. In: *4th International Conference on Learning Representations (ICLR 2016)*.

Pinto J and Fern A (2017) Learning partial policies to speedup MDP tree search via reduction to I.I.D. learning. *Journal of Machine Learning Research* 18(65): 1–35.

Scarselli F, Gori M, Tsoi AC, Hagenbuchner M and Monfardini G (2009) The graph neural network model. *IEEE Transactions on Neural Networks* 20(1): 61–80.

Schulman J, Duan Y, Ho J, et al. (2014) Motion planning with sequential convex optimization and convex collision checking. *The International Journal of Robotics Research* 33(9): 1251–1270.

Silver D, Huang A, Maddison C, et al. (2016) Mastering the game of Go with deep neural networks and tree search. *Nature* 529: 484–489.

Silver D, Schrittwieser J, Simonyan K, et al. (2017) Mastering the game of Go without human knowledge. *Nature* 550: 354–359.

Sperduti A and Starita A (1997) Supervised neural networks for the classification of structures. *IEEE Transactions on Neural Networks* 8(3): 714–735.

Srivastava S, Fang E, Riano L, Chitnis R, Russell S and Abbeel P (2014) Gaussian Process optimization in the bandit setting: no regret and experimental design. In: *IEEE International Conference on Robotics and Automation*.

Stilman M and Kuffner JJ (2005) Navigation among movable obstacles: Real-time reasoning in complex environments. *The International Journal of Humanoid Robotics* 2(4): 479–503.

Stilman M, Schamburek JU, Kuffner J and Asfour T (2007) Manipulation planning among movable obstacles. In: *IEEE International Conference on Robotics and Automation*.

Sutton R and Barto AG (1998) *Reinforcement Learning: An Introduction*. Cambridge, MA: MIT Press.

Toussaint M (2015) Logic-geometric programming: An optimization-based approach to combined task and motion planning. In: *International Joint Conference on Artificial Intelligence*.

Tsochantaridis I, Joachims T, Hofmann T, Altun Y and Singer Y (2005) Large margin methods for structured and interdependent output variables. *Journal of Machine Learning Research* 6(50): 1453–1484.

Wu Z, Pan S, Chen F, Long G, Zhang C and Yu PS (2019) A comprehensive survey on graph neural networks. *arXiv preprint arXiv:1901.00596*.

Yoon SW, Fern A and Givan R (2006) Learning heuristic functions from relaxed plans. In: *International Conference on Automated Planning and Scheduling*.

Zhou J, Cui G, Zhang Z, Yang C, Liu Z and Sun M (2018) Graph neural networks: A review of methods and applications. *arXiv preprint arXiv:1812.08434*.

Zucker M, Ratliff N, Dragan A, et al. (2013) CHOMP: Covariant Hamiltonian optimization for motion planning. *The International Journal of Robotics Research* 32(9–10): 1164–1193.

# Appendix

## A.1. Probabilistically complete version of *SAHS*

Algorithm 4 describes probabilistically complete version of SAHS, PC-SAHS, which takes as inputs an initial state $s_0$, set of goal states $\mathcal{G}$, initial planning horizon $L$, hyperparameters for sampling continuous parameters of operators $N_{\text{fc}}$, $N_{\text{mp}}$, and $N_{\text{smpl}}$, and an abstract-edge heuristic function $\mathbf{h}(\cdot, \cdot)$. We assume we use a probabilistically motion planner for each operator.

The algorithm proceeds similarly to SAHS, but the crucial difference is in line 17 and line 10. What line 17 effectively does is considering all abstract actions sequence of length $L \times 2^{iter}$ at each iteration. For each iteration, we can use at most $N_{\text{fc}} \times 2^{iter}$ number of samples for motion planning, which is enforced by line 10.

We have the following theorem for PC-SAHS.

---

**Algorithm 4** PC-SAHS($s_0, \mathcal{G}, L, N_{\text{fc}}, N_{\text{smpl}}, N_{\text{mp}}, \mathbf{h}(\cdot, \cdot)$)

---

1:   $iter = 0$
2:   **while** not budget_reached
3:     $queue = \text{PriorityQueue}()$
4:     **for** $\delta \in \mathbf{O}^{(M)} \times \mathbf{R}, \mathfrak{a} \in \mathcal{A}$
5:       $queue.add((s_0, \mathfrak{a}(\delta)), \mathbf{h}(\alpha(s, \mathcal{G}), \mathfrak{a}(\delta)))$
6:     **end for**
7:     **while** not $queue.$empty
8:       $s, \mathfrak{a}(\delta) = queue.pop()$
9:       $\kappa = \text{SMPLCONT}(s, \mathfrak{a}(\delta), N_{\text{fc}} \times 2^{iter}, N_{\text{smpl}}, N_{\text{mp}})$
10:      **if** $\kappa$ is feasible
11:       $s' = T(s, \mathfrak{a}(\delta, \kappa))$
12:       $s'.plan = s.plan + \mathfrak{a}(\delta, \kappa)$
13:       **if** $s' \in \mathcal{G}$
14:         **return** $s'.plan$
15:       **end if**
16:       **if** $s'.plan.length < L \times 2^{iter}$
17:        **for** $\delta \in \mathbf{O}^{(M)} \times \mathbf{R}, \mathfrak{a} \in \mathcal{A}$
18:         $queue.add((s', \mathfrak{a}(\delta)), \mathbf{h}(\alpha(s, \mathcal{G}), \mathfrak{a}(\delta)))$
19:       **end for**
20:       **end if**
21:      **end if**
22:     **end while**
23:     $iter = iter + 1$
24:   **end while**

---

**Theorem 1.** *Assume that the motion planner associated with each operator is probabilistically complete. Then, pc-SAHS is probabilistically complete.*

To prove this, first define $B_r(\kappa)$ as a ball with radius $r$ centered at $\kappa$. We make the following assumptions.

**A 1.** *There exists some feasible plan* $[\mathfrak{a}(\delta_1^*, \kappa_1^*), \mathfrak{a}(\delta_2^*, \kappa_2^*), \ldots, \mathfrak{a}(\delta_k^*, \kappa_k^*)]$ *and some* $\varepsilon > 0$ *such that for any* $\hat{\kappa}_1 \in B_\varepsilon(\kappa_1^*), \hat{\kappa}_2 \in B_\varepsilon(\kappa_2^*), \ldots, \hat{\kappa}_k \in B_\varepsilon(\kappa_k^*),$ $[\mathfrak{a}(\delta_1^*, \hat{\kappa}_1), \mathfrak{a}(\delta_2^*, \hat{\kappa}_2), \ldots, \mathfrak{a}(\delta_k^*, \hat{\kappa}_k)]$ *is also a feasible plan, and for each operator* $\mathfrak{a}(\delta_i^*, \hat{\kappa}_i)$ *within such a plan, there exists an IK solution and motion plan that has at least $\varepsilon$ clearance from every obstacle.*

**A 2.** *The underlying feasibility checker (i.e., the IK solver and motion planner) is probabilistically complete, given the above assumption.*

**Lemma 2.** *For some constant c, the probability that we find a solution approaches* 1 *as the number of outer-loop iterations where* $L \times 2^{iter} \geq c$ *and* $N_{\text{fc}} \times 2^{iter} \geq c$ *approaches* $\infty$.

*Proof.* We define an *attempt* for a plan skeleton $\{\delta_1, \delta_2, \ldots, \delta_k\}$ as a sequence of calls to SMPLCONT for each $\delta_i$, with each call starting from the state resulting from the previous call, and terminating if any such call fails. An attempt can occur asynchronously (i.e., other calls to SMPLCONT can take place between those that make up any given attempt), and its calls to SMPLCONT can overlap with attempts for other plan skeletons. Then each outer-loop iteration attempts every plan skeleton of length at most $L \times 2^{iter}$. Therefore, the number of times any plan skeleton of any finite length is attempted approaches $\infty$ as the number of outer-loop iterations where $L \times 2^{iter} \geq c$ approaches $\infty$. In particular, there exists a plan skeleton that feasibly reaches the goal that is attempted arbitrarily many times.

Consider a single attempt for this plan skeleton. Given that it is feasible, then there exists some constant $c$ such that a single call to SMPLCONT succeeds with some positive finite probability when $N_{\text{fc}} \times 2^{iter} \geq c$, by our assumption that the motion planner is probabilistically complete. Then each attempt also succeeds with some positive finite probability because it is the conjunction of the individual calls to SMPLCONT. As each attempt is an independent event, then the probability that at least 1 attempt succeeds approaches 1 as the number of attempts approaches $\infty$. $\square$

**Lemma 3.** *For any constant c, the number of outer-loop iterations where* $L \times 2^{iter} \geq c$ *approaches* $\infty$ *as time approaches* $\infty$.

*Proof.* Suppose it is not. Then because each iteration increases $L$, this means that there is some iteration that does not end in finite time, and so it is stuck in the inner loop forever. Each iteration of the inner loop ends in finite time for a given $N_{\text{smpl}}$ and $N_{\text{mp}}$, so this implies there are an infinite number of iterations in the inner loop. However, each iteration of the inner loop can be mapped to a distinct plan skeleton of length at most $L \times 2^{iter}$, of which there are finitely many for a given value of $L$ and *iter*. Therefore, there are a finite number of inner-loop iterations, and so we have a contradiction. $\square$

Probabilistic completeness follows as a corollary of these two lemmas.

## A.2. Caching the state computation

Although the techniques described here are tied to our specific implementation, the approach is quite general.

### A.2.1. Probabilistic roadmap for predicate computation

To evaluate the predicates PREFREE, MANIPFREE, OCCLUDESPRE, and OCCLUDESMANIP, we use random sampling to sample continuous actions for each object and region and call a motion planner. In our environment, the fixed obstacles remain constant across all problem instances, as only the movable objects have varying initial poses. Therefore, we pre-compute a finely sampled probabilistic roadmap (PRM) that ignores movable objects but respects fixed objects. Later, when doing the graph search for a path, we check for collisions for motions on the edges of the PRM against only the movable objects but not the fixed obstacles in the state. This leads to more efficient and less variable motion planning calls.

### A.2.2. Cached collisions and paths

In a single state we make many motion planning calls. Performing collision checks between movable objects and robot configurations for each motion planning call can be quite expensive. Thus, we cache which configurations in the PRM collide with each object in the current state, then reuse that information in future graph searches as long as that object is not moved. We also retain collision-free paths that are reused in multiple predicate evaluations.

### A.2.3. IK solutions

Unlike in the box-moving domain, in which collisions mostly constrain the space of feasible trajectories of the robot base, collisions in the cupboard domain heavily constrain the space of feasible arm configurations instead. Therefore, many IK solutions are in collision, and so we must sample many configurations in order to find a feasible operator instance. Because IK solving is a relatively expensive operation, this severely affects the efficiency of planning and evaluating predicates. The workaround we use is to pre-compute a large number of IK solutions for objects at a wide variety of poses relative to the robot base. Then when computing the geometric predicates at planning time we adapt the cached configuration by using relative transformations to make it fit with the actual object pose (for a pick operation) or the desired object placement (for a place operation). Many of these cached solutions will still be in collision, but avoiding the cost of finding the kinematic solution leads to significant speedups. We still call the IK solver when sampling continuous parameters after selecting an abstract action to attempt.

### A.2.4. Predicate evaluations

Finally, when an action is applied to a state, resulting in a new state, a lot of information can be passed down to improve the efficiency of evaluating the predicates for the new state. First, for any given action, many predicates will not change because moving a single object will leave most relationships between other objects the same, so we reuse the predicate value without recomputing it. Additionally, the set of PRM configurations in collision with each object only changes for the object that was moved, and so all other sets of collisions can be reused. In both domains, IK solutions that are known to not collide with fixed objects can be retained for all objects except for the one that was moved. These configurations might still collide with movable objects, though.

## A.3. Computations inside our rank function GNN

We now describe in detail the computations inside our GNN. For each operator $\alpha$, such as PICKANDPLACE, we define a GNN $\widehat{F}_\alpha(\alpha(s, \mathcal{G}), \delta)$ that ranks discrete parameter choices for all $\delta \in \mathbf{O}^{(M)} \times \mathbf{R}$. Each GNN takes the graph input such as that in Figure 6 as an input, and outputs a $|\mathbf{O}^{(M)}|$ by $|\mathbf{R}|$ matrix where each value in the matrix indicates the rank value of moving an object to a region.

Because the arguments of our edge function Edge($\cdot$, $\cdot$, $\cdot$) are ordered, we call the entity in the first argument as a *sender*, and that in the second argument as a *receiver*. We found that we achieve better performance if we use a separate embedding functions for the sender and a receiver. We first embed the values at the object nodes as

$$u_{o_i}^{(0)} = f(\text{Node}(o_i), \theta_1) \quad \text{and} \quad v_{o_j}^{(0)} = f(\text{Node}(o_j), \theta_2)$$

for all $o_i, o_j \in \mathbf{O}^{(M)}$, where the superscript denotes the round of message passing, $u$ denotes the embedding of the sender $o_i$, and $v$ denotes the embedding of the receiver $o_j$.

At each edge, we compute an edge embedding with

$$c_{o_i o_j r_k} = f(\text{Edge}(o_i, o_j, r_k); \theta_3)$$

We compute the message from $o_i$ to $o_j$ for the component associated region $r_k$, $m_{o_i o_j r_k}$, by using these sender, receiver, and edge embeddings,

$$m_{o_i o_j r_k}^{(0)} = f(u_{o_i}^{(0)}, v_{o_j}^{(0)}, c_{o_i o_j r_k}; \theta_4).$$

We use the standard averaging-aggregation function to aggregate messages. However, one key distinction is that although we do not have an edge between components, we aggregate messages across components, because the occlusion information in both regions are necessary to predict the rank value of moving an object to a region. Thus, we aggregate by

$$m_{o_j}^{(0)} = \frac{1}{|\mathbf{O}^{(M)}| + |\mathbf{R}|} \sum_{o_i, r_k} m_{o_i o_j r_k}^{(0)}$$

We perform one more round of message passing. The values at nodes are computed again using $f(\cdot; \theta_1)$ and $f(\cdot; \theta_2)$ as

$$u_{o_j}^{(1)} = f(m_{o_j}^{(0)}; \theta_1) \quad \text{and} \quad v_{o_j}^{(1)} = f(m_{o_j}^{(0)}; \theta_2)$$

The new messages are computed with the updated node values and the edge embedding

$$m_{o_i o_j r_k}^{(1)} = f(u_{o_j r_k}^{(1)}, v_{o_j r_k}^{(1)}, c_{o_i o_j r_k}, \theta_4)$$

We then aggregate the messages using averaging, only with respect to objects this time, to have distinct values for different regions for $(o, r)$ pairs:

$$m_{o_j r_k}^{(1)} = \frac{1}{|\mathbf{O}^{(M)}|} \sum_{o_i} m_{o_i o_j r_k}^{(0)}$$

Finally, we compute the rank value of moving $o_j$ to region $r_k$ in an abstract state $\alpha(s, \mathcal{G})$ as a matrix with size $|\mathbf{O}^{(M)}| \times |\mathbf{R}|$, where each entry is

$$\widehat{F}_\alpha(\alpha(s, \mathcal{G}), (o_j, r_k); \theta) = f(m_{o_j, r_k}^{(1)}; \theta_5)$$

where $\theta = \{\theta_1, \theta_2, \theta_3, \theta_4, \theta_5\}$

## A.4. Cleaning the training dataset

As our planning algorithm uses sampling to handle continuous parameters, the continuous parameter data tends to noisy. For example, we often get suboptimal state–continuous-parameter pairs, where the robot places an object at a superfluous pose only to move it again later. If we use this dataset directly, then we would end up with a sampler that is very similar to a uniform sampler, because there is not enough regularity between states and parameters.

To deal with this problem, we use a strategy for cleaning the dataset such that each parameter is a progress towards a goal. The idea is to check whether the object that we move at each step decreases the number of objects in collision with $V_{\text{pre}}$ or $V_{\text{manip}}$ for goal objects into their associated goal regions.

To build such dataset, we first consider a tuple $(s_t, o_t, \kappa_t, s_{t+1})$ from $\mathbf{D}_K$, where $o_t$ is the object that the robot moves at time step $t$, $\kappa_t$ is the continuous parameter that moves $o_t$, and $s_{t+1}$ is the resulting next state. We wish to determine whether to include $\kappa_t$ into our training data $\mathbf{D}_K$ or not.

To do this, we first look at, for each $o_g \in \mathcal{G}$, the number of times $o_t$ is in collision with $V_{\text{pre}}(o_g)$ at $s_t$. Then, for each $(o_g, r_g) \in \mathcal{G}$, we look at the number of times $o_t$ is in collision with $V_{\mathrm{manip}}(o_g, r_g)$. We count the number of collisions, and denote it as $m_t$. We repeat these steps in state $s_{t+1}$, and compute $m_{t+1}$. We include $\kappa_t$ in $\mathbf{D}_K$ only if $m_{t+1} - m_t > 0$.

Intuitively, this method makes sure that each parameter we include in our data moves an object out of swept volumes that needs to be cleared to achieve a goal. This creates a regularity between a state and parameter especially because $\phi(s)$ encodes $V_{\mathrm{manip}}$ s for all the goal object and goal region pairs. Note, however, that in very tight environments where we need to first place objects into these essential swept volumes, this method will discard data rather too aggressively.

## A.5. Hyperparameters and training data description

For the parameters of SAHS, we use $N_{smpl} = 2,000$, $N_{mp} = 5$ in the box-moving domain, and $N_{smpl} = 50$ for the shelf domain. The priority function in 2 uses $\lambda = 1$. For training the sampler, we use $n_{tot} = 100,000$, $n_c = 5$, $n_b = 32$, $lr_\theta = 1 \times 10^{-4}$, $lr_\alpha = 1 \times 10^{-4}$, and use $\mathcal{N}(0, 1)$ for $P_Z$.

We train separate samplers for pick parameters and place parameters, and feed the predicted pick parameters to the place parameter sampler. We also train separate samplers for different regions. We use 1,000 planning episodes for training samplers in the cupboard domain, and 1,500 planning episodes for training samplers in the box-moving domain. For training the ranking function, we use 250 planning episodes.

To generate our training data, we first solve problems using IRSC and use the planning experience to train our ranking function. Then, we use SAHS-RANK to solve additional problems and use this planning experience to train our sampler. To build the set of key configurations, we use the motion plans from IRSC planning experience, and then sparsely subsample them by discarding those that are too close.

The learned samplers, which operates on lower geometric details, however, must be trained in each environment. For this reason, its planning experience data consists of a mixture of moving 1 and 4 boxes into the kitchen in the box-moving domain, and packing 1 object in the cupboard domain. We train separate samplers for these environments.